\documentclass{article} % For LaTeX2e
\usepackage{iclr2021_conference,times}

\usepackage{xfrac}
\usepackage[T1]{fontenc}    % use 8-bit T1 fonts
\usepackage{hyperref}       % hyperlinks
\usepackage{url}            % simple URL typesetting
\usepackage{booktabs}       % professional-quality tables
\usepackage{amsfonts}       % blackboard math symbols
\usepackage{nicefrac}       % compact symbols for 1/2, etc.
\usepackage{microtype}      % microtypography
\usepackage[algoruled,boxed,lined]{algorithm2e}
\usepackage{enumitem}
\usepackage{graphicx}

\usepackage{amsmath}
\usepackage{amssymb}
\usepackage{mathtools}

\usepackage{lineno}
\usepackage{amsthm}
\usepackage{xcolor}

\newtheorem{theorem}{Theorem}[section]

\DeclareMathOperator*{\argmax}{arg\,max}
\DeclareMathOperator*{\argmin}{arg\,min}

\DeclareMathOperator{\Tr}{Tr}

\usepackage{amssymb}
\usepackage{appendix}

\usepackage{caption}
\usepackage{subcaption}
\usepackage{wrapfig}
\usepackage{bm}
\usepackage{multirow}
\iclrfinalcopy

\allowdisplaybreaks
\setlist[itemize]{noitemsep, nolistsep, leftmargin=*}
\setlist[enumerate]{noitemsep, nolistsep, leftmargin=*}

\title{Continuous Wasserstein-2 Barycenter\\Estimation without Minimax Optimization}

\author{Alexander Korotin\\
Skolkovo Institute of Science and Technology\\
\textit{Advanced Data Analytics in Science and}\\ \textit{Engineering Group}\\
Moscow, Russia \\
\texttt{a.korotin@skoltech.ru} \\
\And
Lingxiao Li \\
Massachusetts Institute of Technology \\
\textit{Geometric Data Processing Group}\\
Cambridge, Massachusetts, USA \\
\texttt{lingxiao@mit.edu} \\
\And
Justin Solomon \\
Massachusetts Institute of Technology \\
\textit{Geometric Data Processing Group}\\
Cambridge, Massachusetts, USA \\
\texttt{jsolomon@mit.edu}\hspace{43mm}
\And
Evgeny Burnaev \\
Skolkovo Institute of Science and Technology\\
\textit{Advanced Data Analytics in Science}\\
\textit{and Engineering Group}\\
Moscow, Russia \\
\texttt{e.burnaev@skoltech.ru} \\
}

\begin{document}

\maketitle

\begin{abstract}
Wasserstein barycenters provide a geometric notion of the weighted average of probability measures based on optimal transport. In this paper, we present a scalable algorithm to compute Wasserstein-2 barycenters given sample access to the input measures, which are not restricted to being discrete. While past approaches rely on entropic or quadratic regularization, we employ input convex neural networks and cycle-consistency regularization to avoid introducing bias. As a result, our approach does not resort to minimax optimization. We provide theoretical analysis on error bounds as well as empirical evidence of the effectiveness of the proposed approach in low-dimensional qualitative scenarios and high-dimensional quantitative experiments.
\end{abstract}
% \lingxiao{should explain what ``continuous'' here means. Emphasize that both the inputs and the output are continuous (e.g. allow sample access).
% Maybe also emphasize there is no entropic/L2 regularization}. 
% Our method builds on input convex neural networks and cycle-consistency regularization for computing the Wasserstein-2 distance. 
% \justin{this doesn't seem like the right way to use the term ``end-to-end''; I think minimax methods can be end-to-end so long as they aren't trained in stages} 
% Our method builds on input convex neural networks with . % for computing the Wasserstein-2 distance. 

\section{Introduction}

% averaging suitable sense, interpolation
% Averaging probability distributions is a 

% The Wasserstein-2 barycenter is a geometrically meaningful notion of the weighted average of a population of probability measures.
% \lingxiao{don't repeat abstract}
Wasserstein barycenters have become popular due to their ability to represent the average of probability measures in a geometrically meaningful way.
% \lingxiao{this sentence is repeating what the previous sentence said}. 
Techniques for computing Wasserstein barycenters have been successfully applied to many computational problems.
% \lingxiao{I think this sentence and the rest of the paragraph should be merged with the first paragraph. This sentence is also kinda repeating the previous one.}.
In image processing, Wasserstein barycenters are used for color and style transfer \citep{rabin2014adaptive,mroueh2019wasserstein}, and texture synthesis \citep{rabin2011wasserstein}.
% \lingxiao{don't use etc.; either list more or don't}. 
In geometry processing, shape interpolation can be done by computing barycenters \citep{solomon2015convolutional}. 
In online machine learning, barycenters are used for aggregating probabilistic predictions of experts \citep{korotin2019integral}. 
Within the context of Bayesian inference, the barycenter of subset posteriors converges to the full data posterior, thus enabling efficient computational methods based on finding the barycenters \citep{srivastava2015wasp,srivastava2018scalable}.

% ,li2020continuous}.
% \lingxiao{This paragraph could use some better transitions, instead of just listing ``In XXX, .... In XXX, ....''}

% Thus, Wasserstein-2 barycenter provides the weighted averaging of probability measures in a suitable sense.
% behaving a lot similar to averaging of vectors in $\mathcal{L}^{2}$.

Fast and accurate barycenter algorithms exist for discrete distributions (see \cite{peyre2019computational} for a survey), while for continuous distributions the situation is more difficult and remains unexplored until recently \citep{li2020continuous, fan2020scalable, cohen2020estimating}.
The discrete methods scale poorly with the number of support points of the barycenter and thus cannot approximate continuous barycenters well, especially in high dimensions.
% \lingxiao{the previous sentence says fast, but here it contradicts ``'do not scale well'}. 
% due to curse of dimensionality, \lingxiao{``curse of dimensionality'' seems unfit here, consider removing}
%As a result,
%they cannot be efficiently applied to approximate barycenters of continuous measures in high dimensions.
% \lingxiao{add ``in higher dimensions''}.

In this paper, we present a method to compute Wasserstein-2 barycenters of \emph{continuous} distributions based on a novel regularized dual formulation where the convex potentials are parameterized by input convex neural networks \citep{amos2017input}.
Our algorithm is straightforward without introducing bias (e.g. \cite{li2020continuous}) or requiring minimax optimization (e.g. \cite{fan2020scalable}).
%In this paper, we propose a
% \justin{again I don't know of algorithms for W2 barycenters that \emph{aren't} end-to-end}
%non-minimax algorithm for computation of Wasserstein-2 barycenters in the continuous case. 
% \justin{Say more.  Summarize your algorithm in 2-3 sentences.  Currently your introduction sets up the problem but doesn't say how we're going to solve it!}
% Our approach relies on a novel regularized dual Wasserstein-2 barycenter problem. 
This is made possible by combining a new \emph{congruence regularizing term} combined with \emph{cycle-consistency regularization} \citep{korotin2019wasserstein}.
As we will show in the analysis, thanks to the properties of Wasserstein-2 distances, the gradients of the resulting convex potentials ``push'' the input distributions close to the true barycenter, allowing good approximation of the barycenter.

% \textbf{The article is structured as follows}. In Section \ref{sec-preliminaries}, we give preliminary notions on Wasserstein-2 distance and Wasserstein-2 barycenters. In Section \ref{sec-related-work}, we discuss related work. In Section \ref{sec-algorithm}, we analytically derive our algorithm and provide performance guarantees. The numerical optimization procedure of the algorithm is given in Appendix \ref{sec-algorithm-procedure}. The proofs of theoretical results are given in Appendix \ref{sec-proofs}. In Section \ref{sec-experiments}, we evaluate the method and compare it with existing alternatives. In Appendix \ref{sec-exp-toy-extra}, we provide additional experiments and list the hyperparameters we used. % justin commented this out.  it just echoes the paper text, and the template here is too small for redundancy

\section{Preliminaries}
\label{sec-preliminaries}
We denote the set of all Borel probability measures on $\mathbb{R}^{D}$ with finite second moment by $\mathcal{P}_{2}(\mathbb{R}^{D})$. We use $\mathcal{P}_{2,\text{ac}}(\mathbb{R}^{D})\subset \mathcal{P}_{2}(\mathbb{R}^{D})$ to denote the subset of all absolutely continuous measures (w.r.t. the Lebesgue measure).
% $\mathbb{P}_{1},\dots,\mathbb{P}_{K}$ are probability distributions over the Borel field of $\mathbb{R}^{D}\subset\mathbb{R}^{D}$ with finite second moments.

\paragraph*{Wasserstein-2 distance. }
For $\mathbb{P},\mathbb{Q}\in\mathcal{P}_{2}(\mathbb{R}^{D})$, %(the square of) 
the \textbf{Wasserstein-2} distance is defined by
\begin{equation}\mathbb{W}_{2}^{2}(\mathbb{P},\mathbb{Q})\stackrel{\text{def}}{=}\min_{\pi\in\Pi(\mathbb{P},\mathbb{Q})}\int_{\mathbb{R}^{D}\times \mathbb{R}^{D}}\frac{\|x-y\|^{2}}{2}d\pi(x,y),
\label{w2-primal-form}
\end{equation}
% \lingxiao{should we use $\inf$ instead of $\min$ to be precise? $\min$ is reserved for optimization over a finite set}
where $\Pi(\mathbb{P},\mathbb{Q})$ is the set of probability measures on $\mathbb{R}^{D}\times\mathbb{R}^{D}$ whose marginals are $\mathbb{P},\mathbb{Q}$, respectively.
This definition is known as Kantorovich's primal form of transport distance \citep{kantorovitch1958translocation}. 

The Wasserstein-2 distance $\mathbb{W}_{2}$ is well-studied in the theory of optimal transport \citep{brenier1991polar,mccann1995existence}. In particular, it has a \textbf{dual formulation} \citep{villani2003topics}:
\begin{eqnarray}\mathbb{W}_{2}^{2}(\mathbb{P}, \mathbb{Q})=\int_{\mathbb{R}^{D}}\frac{\|x\|^{2}}{2}d\mathbb{P}(x)+\int_{\mathbb{R}^{D}}\frac{\|y\|^{2}}{2}d\mathbb{Q}(y)-
\min_{\psi\in\text{Conv}}\bigg[\int_{\mathbb{R}^{D}}\psi(x)d\mathbb{P}(x)+\int_{\mathbb{R}^{D}}\overline{\psi}(y)d\mathbb{Q}(y)\bigg],
\label{w2-dual-form}
\end{eqnarray}
where the minimum is taken over all the convex functions (potentials) $\psi:\mathbb{R}^{D}\rightarrow \mathbb{R}\cup \{\infty\}$, and
% \begin{equation}
$\overline{\psi}(y)=\max_{x\in\mathbb{R}^{D}}\big(\langle x, y\rangle - \psi(x)\big):\mathbb{R}^{D}\rightarrow \mathbb{R}\cup \{\infty\}$
% \label{conv-conjugate}
% \end{equation}
is the \textbf{convex conjugate} of $\psi$ \citep{fenchel1949conjugate}, which is also a convex function. 
The optimal potential $\psi^{*}$ is defined up to an additive constant.

\cite{brenier1991polar} shows that
% \justin{replace with original citation to Brenier?} theorem \cite[Theorem 2.12]{villani2003topics}
if $\mathbb{P}$ does not give mass to sets of dimensions at most $D-1$, then the optimal plan $\pi$ is uniquely determined by $\pi^{*}=[\text{id}_{\mathbb{R}^{D}}, T^{*}]\sharp \mathbb{P}$, where $T^*:\mathbb{R}^{D}\rightarrow\mathbb{R}^{D}$ is the unique solution to the Monge's problem
\begin{equation}
T^{*}=\argmin_{T\sharp\mathbb{P}=\mathbb{Q}}\int_{\mathbb{R}^{D}}\frac{\|x-T(x)\|^{2}}{2}d\mathbb{P}(x).
\label{w2-monge-form}
\end{equation}
The connection between $T^{*}$ and the dual formulation \eqref{w2-dual-form} is that $T^{*}=\nabla \psi^{*}$, where $\psi^{*}$ is the optimal solution %(convex potential) 
of \eqref{w2-dual-form}. 
% \lingxiao{should change the reference style to Equation 2}
Additionally, if $\mathbb{Q}$ does not give mass to sets of dimensions at most $D-1$, 
% \lingxiao{small sets = D-1 dimensional sets?}
then $T^*$ is invertible and
$$T^{*}(x)=\nabla \psi^{*}(x)=(\nabla\overline{\psi^{*}})^{-1}(x),\qquad (T^{*})^{-1}(y)=\nabla\overline{\psi^{*}}(y)=(\nabla \psi^{*})^{-1}(y).$$
In particular, the above discussion applies to the case where $\mathbb{P},\mathbb{Q}\in \mathcal{P}_{2,ac}(\mathbb{R}^{D})$.

\paragraph*{Wasserstein-2 barycenter. }
Let $\mathbb{P}_{1},\dots,\mathbb{P}_{N}\in \mathcal{P}_{2,ac}(\mathbb{R}^{D})$.
% \lingxiao{"population" may not be accurate here. I would think "population" as an infinite set}
Then, their barycenter w.r.t.\ weights $\alpha_{1},\dots,\alpha_{N}$ ($\alpha_{n}> 0$ and $\sum_{n=1}^{N}\alpha_{n}=1$) is %defined by
\begin{equation}
\overline{\mathbb{P}}\stackrel{\text{def}}{=}\argmin_{\mathbb{P}\in\mathcal{P}_{2}(\mathbb{R}^{D})}\sum_{n=1}^{N}\alpha_{n}\mathbb{W}_{2}^{2}(\mathbb{P}_{n},\mathbb{P}).
\label{w2-barycenter-def}
\end{equation}
% The functional $\text{BF}(\mathbb{P})$ appearing inside the argmin is called the \textbf{barycenter functional}. \lingxiao{The notation BF doesn't seem necessary and is inaccurate since it depends on all the input}
%
Throughout this paper, we assume that at least one of ${\mathbb{P}_{1},\dots,\mathbb{P}_{N}\in \mathcal{P}_{2,ac}(\mathbb{R}^{D})}$ has bounded density. Under this assumption, $\overline{\mathbb{P}}$ is unique and absolutely continuous, i.e.,  $\overline{\mathbb{P}}\in\mathcal{P}_{2,ac}(\mathbb{R}^{D})$, and it has bounded density \citep[Definition 3.6 \& Theorem 5.1]{agueh2011barycenters}.

For $n\in\{1,2,\dots,N\}$, let $(\psi^{*}_{n},\overline{\psi^{*}_{n}})$ be the optimal pair of (mutually) conjugate potentials that transport $\mathbb{P}_{n}$ to $\overline{\mathbb{P}}$, i.e., $\nabla\psi^{*}_{n}\sharp\mathbb{P}_{n}=\overline{\mathbb{P}}$ and $\nabla\overline{\psi^{*}_{n}}\sharp\overline{\mathbb{P}}=\mathbb{P}_{n}$. %According to the characterization of barycenters  , 
Then $\{\overline{\psi^{*}_{n}}\}$ satisfy
\begin{equation}\sum_{n=1}^{N}\alpha_{n}\nabla\overline{\psi^{*}_{n}}(x)=x\qquad \text{and}\qquad  \sum_{n=1}^{N}\alpha_{n}\overline{\psi^{*}_{n}}(x)=\frac{\|x\|^{2}}{2}+c.
\label{bar-char}
\end{equation}
% \justin{picky comment:  in the equation above, should you replace $\Longleftrightarrow$ with $\textrm{and}$ to avoid reading it as ``if and only if'' ?}
for all $x\in\mathbb{R}^{D}$ 
\citep{agueh2011barycenters,alvarez2016fixed}.
Since optimal potentials are defined up to a constant, for convenience, we set $c=0$. The condition \eqref{bar-char} serves as the basis for our algorithm for computing Wasserstein-2 barycenters. %
We say that potentials $\psi_{1},\dots,\psi_{N}$ are \textbf{congruent} w.r.t.\ weights $\alpha_{1},\dots,\alpha_{n}$ if their conjugate potentials satisfy \eqref{bar-char}, i.e.,  $\sum_{n=1}^{D}\alpha_{n}\overline{\psi_{n}}(x)=\frac{\|x\|^{2}}{2}$ for all $x\in\mathbb{R}^{D}$.
%  \lingxiao{iff doesn't seem formal. "if" seems fine since you are defining a term}

\section{Related Work}
\label{sec-related-work}
Most algorithms in the field of computational optimal transport are designed for the discrete setting where the input distributions have finite support; see the recent survey by \citet{peyre2019computational} for discussion.
A particular popular line of algorithms are based on entropic regularization that gives rise to the famous Sinkhorn iteration \citep{cuturi2013sinkhorn, cuturi2014fast}. 
% In this case, efficient algorithms for computing optimal transport distances and maps exist, see . The discrete barycenter problem is also well-studied, see e.g. algorithms \citep{}.
These methods are typically limited to a support of $10^{5}\!-\!10^{6}$ points before the problem becomes computationally infeasible. 
Similarly, discrete barycenter methods \citep{cuturi2014fast}, particularly the ones that rely on a fixed support for the barycenter \citep{dvurechenskii2018decentralize, staib2017parallel}, cannot provide precise approximation of continuous barycenters in high dimensions, since a large number of samples is needed; see experiments in \citet[\S4.3]{fan2020scalable} for an example.
% \lingxiao{Is ``curse of dimensionality'' accurate here? Might be safer to just say the discrete algorithms are not suited for high dimensions where a large number of samples are needed}
Thus we focus on the existing literature in the continuous setting.
% We do not discuss discrete barycenter algorithms in the paper. \lingxiao{discuss is too strong here. Should instead say: thus, we will focus only on continuous barycenter algorithms}

\paragraph*{Computation of Wasserstein-2 distances and maps. }
\citet{genevay2016stochastic} demonstrate the possibility of computing Wasserstein distances given only sample access to the distributions by parameterizing the dual potentials as functions in the reproducing kernel Hilbert spaces.
Based on this realization, \citet{seguy2017large} propose a similar method but use neural networks to parameterize the potentials, using entropic or $\mathcal L^2$ regularization w.r.t.\ $\mathbb{P}\times\mathbb{Q}$ to keep the potentials approximately conjugate.  
The transport map is recovered from optimized potentials via barycentric projection.
% \lingxiao{This is not correct. In seguy's paper their preferred way is to train a separate neural network for estimating the map, not barycentric projection}.

As we note in \S\ref{sec-preliminaries},
% \lingxiao{Subsection is weird. Why not just Section? Also you can just use autoref or cref to automatically inject Section}
$\mathbb{W}_{2}$ enjoys many useful theoretical properties. 
For example, the optimal potential $\psi^{*}$ is convex, and the corresponding optimal transport map is given by $\nabla\psi^{*}$. By exploiting these properties, \citet{makkuva2019optimal} propose a minimax optimization algorithm for recovering transport maps, using input convex neural networks (ICNNs) \citep{amos2017input} to approximate the potentials.

An alternative to entropic regularization is the \textbf{cycle-consistency} regularization proposed by \citet{korotin2019wasserstein}.
% Analogously to \citep{seguy2017large}, the method approximates the potentials in \eqref{w2-dual-form} with neural networks. 
It uses the property that the gradients of optimal dual potentials are inverses of each other. 
The imposed regularizer requires integration only over the marginal measures $\mathbb{P}$ and $\mathbb{Q}$, instead of over $\mathbb{P}\times\mathbb{Q}$ as required by entropy-based alternatives. 
% Such method scales better with the dimension than the entropy-based alternative. 
Their method converges faster than the minimax method since it does not have an inner optimization cycle. 
% {\color{red}we should add a section to the paper on W2GN!!!}

\cite{xie2019scalable} propose using two generative models with a shared latent space to implicitly compute the optimal transport correspondence between $\mathbb{P}$ and $\mathbb{Q}$. Based on the obtained correspondence, the authors are able to compute the optimal transport distance between the distributions.

\paragraph*{Computation of Wasserstein-2 barycenters. }
% \label{sec-comp-w2}
A few recent techniques tackle the barycenter problem \eqref{w2-barycenter-def} using continuous rather than discrete approximations of the barycenter:
\begin{itemize}
    \item \textsc{Measure-based (generative) optimization:} Problem \eqref{w2-barycenter-def} optimizes over probability measures. This can be done using the generic algorithm by \citet{cohen2020estimating} who employ generative networks to compute barycenters w.r.t.\ arbitrary discrepancies. They test their method with the maximum mean discrepancy (MMD) and Sinkhorn divergence. This approach suffers from the usual limitations of generative models such as mode collapse. Applying it to $\mathbb W_2$ barycenters requires estimation of $\mathbb{W}_{2}^{2}(\mathbb{P}_{n},\mathbb{P})$. \citet{fan2020scalable} test this approach using the minimax method by \citet{makkuva2019optimal}, but they end up with a challenging \emph{min-max-min} problem.
    \vspace{1mm}\item \textsc{Potential-based optimization:} \citet{li2020continuous} recover the optimal potentials $\{\psi^{*}_{n}\}$ via  a non-minimax regularized dual formulation. No generative model is needed: the barycenter is recovered by pushing forward measures using gradients of potentials or by barycentric projection.
\end{itemize}

% \lingxiao{push forward is not a noun. Using push-forward or pushforward instead.}

% Its goal is recover the optimal potentials appearing in the dual formulation of Wasserstein-2 distance $\mathbb{W}_{2}^{2}(\mathbb{P}_{n},\mathbb{Q})$. The barycenter measure is recovered from the obtained potentials (as the barycenter mapping or as a gradient map pushforward measure).

% As far as we know, the only existing paper exploiting this principle is \citep{li2020continuous}. It ...

\section{Methods}
\label{sec-algorithm}

% To begin with, we note that \eqref{w2-barycenter-def} implies optimization over probability measures by using generative networks.
% % \lingxiao{the use of e.g. here is questionable}
% Such approaches naturally suffer from usual problems of generative models, such as mode dropping, mode collapse, disconvergence, etc.

%Our approach is potential-based and does not require a generative model.
% \lingxiao{This sentence is a bit abrupt without any transition}
%We use a potential-based approach rather than a generative model,  
% optimizing over the potentials $\{\psi_{n},\overline{\psi_{n}}\}$ in the dual formulation of Wasserstein distance \eqref{w2-dual-form}. The barycenter is recovered from the gradient of any potential $\nabla \psi_{n}$ as the pushforward measure $\nabla \psi_{n}\sharp\mathbb{P}_{n}$.
%\lingxiao{This paragraph written as such details nothing different from my paper; you should consider either removing this entirely or saying something like ``Similar to \citet{li2020continuous} we use a potential-bawsed method; however the key differences are: (1) we restrict the potentials to be convex (2) our formulation does not introduce bias''}
Inspired by \citet{li2020continuous} we use a potential-based approach and recover the barycenter by using gradients of the potentials as pushforward maps. 
The main differences are: (1) we restrict the potentials to be convex, (2) we enforce congruence via a regularizing term, and (3) our formulation does not introduce bias, meaning the optimal solution of our formulation gives the true barycenter.
%  \justin{isn't the typical notation $(\nabla\psi_n)_\sharp\mathbb P_n$?}
% \lingxiao{mention the usage of pushforward measures}

\subsection{Deriving the Dual Problem}
% To derive our method,
% \lingxiao{gently??}
% we first make an assumption that
% \lingxiao{I think there is no point in mentioning ``cheat''; just say ``let P be the true barycenter''}

Let $\overline{\mathbb{P}}$ be the true barycenter. Our goal is to recover the optimal potentials $\{\psi_{n}^{*},\overline{\psi_{n}^{*}}\}$ mapping the input measures $\mathbb P_n$ into $\overline{\mathbb P}$.

To start, we express the barycenter objective \eqref{w2-barycenter-def} after substituting the dual formulation \eqref{w2-dual-form}:
\begin{eqnarray}
\sum_{n=1}^{N}\alpha_{n}\mathbb{W}_{2}^{2}(\mathbb{P}_{n},\overline{\mathbb{P}})=\bigg[\sum_{n=1}^{N}\alpha_{n}\int_{\mathbb{R}^{D}}\frac{\|x\|^{2}}{2}d\mathbb{P}_{n}(x)\bigg]+\int_{\mathbb{R}^{D}}\frac{\|y\|^{2}}{2}d\overline{\mathbb{P}}(y)-
\nonumber
\\
\min_{\{\psi_{n}\}\in\text{Conv}}\bigg[\sum_{n=1}^{N}\alpha_{n}\int_{\mathbb{R}^{D}}\psi_n(x)d\mathbb{P}_{n}(x)+\sum_{n=1}^{N}\alpha_{n}\int_{\mathbb{R}^{D}}\overline{\psi_{n}}(y)d\overline{\mathbb{P}}(y)\bigg]
\label{bf-w2-dual}
\end{eqnarray}
The minimum is attained not just among convex potentials $\{\psi_{n}\}$, but among congruent potentials (see discussion under \eqref{bar-char}); thus, we can add the constraint that $\{\psi_n\}$ are congruent to \eqref{bf-w2-dual}. Hence,
\begin{eqnarray}
\sum_{n=1}^{N}\alpha_{n}\mathbb{W}_{2}^{2}(\mathbb{P}_{n},\overline{\mathbb{P}})
%=\bigg[\sum_{n=1}^{N}\alpha_{n}\int_{\mathbb{R}^{D}}\frac{\|x\|^{2}}{2}d\mathbb{P}_{n}(x)\bigg]+\int_{\mathbb{R}^{D}}\frac{\|y\|^{2}}{2}d\overline{\mathbb{P}}(y)-
%\nonumber
%\\
%\min_{\{\psi_{n}\}\text{ are Congruent}}\bigg[\sum_{n=1}^{N}\alpha_{n}\int_{\mathbb{R}^{D}}\psi_n(x)d\mathbb{P}_{n}(x)+\sum_{n=1}^{N}\int_{\mathbb{R}^{D}}\alpha_{n}\overline{\psi_{n}}(y)d\overline{\mathbb{P}}(y)\bigg]=
%\label{bf-w2-congruent}
%\\
=\bigg[\sum_{n=1}^{N}\alpha_{n}\int_{\mathbb{R}^{D}}\frac{\|x\|^{2}}{2}d\mathbb{P}_{n}(x)\bigg]-\!\!\!\min_{\{\psi_{n}\}\text{ congruent}}\bigg[\underbrace{\sum_{n=1}^{N}\alpha_{n}\int_{\mathbb{R}^{D}}\!\!\psi_{n}(y)d\mathbb{P}_{n}(y)}_{\text{MultiCorr}(\{\alpha_{n}, \mathbb{P}_{n}\}\vert \{\psi_{n}\})}\bigg].
\label{bf-w2-final}
\end{eqnarray}
% \lingxiao{use a simplified notation $\text{MultiCorr}(\{\psi_n\})$ or $MC(\{\psi_n\})$; you're not changing the weights or the input measures anyway. You are already forgetting $\{\alpha_n\}$ in some places}
To transition from \eqref{bf-w2-dual} to \eqref{bf-w2-final}, we used the fact that for congruent $\{\psi_{n}\}$ we have ${\sum_{n=1}^{N}\alpha_{n}\overline{\psi_{n}}(x)=\frac{\|x\|^{2}}{2}}$, so 
$\sum_{n=1}^{N}\int_{\mathbb{R}^{D}}\alpha_{n}\overline{\psi_{n}}(y)d\overline{\mathbb{P}}(y)=\int_{\mathbb{R}^{D}}\frac{\|y\|^{2}}{2}d\overline{\mathbb{P}}(y).$

We call the value inside the minimum in \eqref{bf-w2-final} the \textbf{multiple correlation} of $\{\mathbb{P}_{n}\}$ with weights $\{\alpha_{n}\}$ w.r.t.\ potentials $\{\psi_{n}\}$.
Notice that the true barycenter $\overline{\mathbb{P}}$ appears nowhere on the right side of \eqref{bf-w2-final}. % and its first term is constant w.r.t. the unknowns.
Thus the optimal potentials $\{\psi_{n}^{*}\}$ can be recovered by solving the following
\begin{eqnarray}
\min_{\{\psi_{n}\}\text{ congruent}}\text{MultiCorr}(\{\alpha_{n}, \mathbb{P}_{n}\}\vert \{\psi_{n}\}) =  \min_{\{\psi_{n}\}\text{ congruent}}\bigg[\sum_{n=1}^{N}\alpha_{n}\int_{\mathbb{R}^{D}}\!\!\psi_{n}(y)d\mathbb{P}_{n}(y)\bigg].
\label{unreg-objective}
\end{eqnarray}
%optimizing multiple correlation over the congruent convex potentials, since the first term of \eqref{bf-w2-final} is constant w.r.t.\ the unknowns.

\subsection{Imposing the Congruence Condition}
It is challenging to impose the congruence condition on convex potentials. 
What if we relax the congruence condition? 
The following theorem bounds how close a set of convex potentials $\{\psi_n\}$ is to $\{\psi_n^*\}$ in terms of the difference of multiple correlation.
%To begin to do so, we consider the case when the potentials are not necessarily congruent, yielding the following theorem:
\begin{theorem} %[Generative Property for Approximators of Multiple Correlations]%<-- name didn't seem useful and takes lots of space
Let $\overline{\mathbb{P}}\in\mathcal{P}_{2,ac}(\mathbb{R}^{D})$ be the barycenter of $\mathbb{P}_{1},\dots,\mathbb{P}_{N}\in\mathcal{P}_{2,ac}(\mathbb{R}^{D})$ w.r.t.\ weights $\alpha_{1},\dots,\alpha_{N}$. Let $\{\psi_{n}^{*}\}$ be the optimal congruent potentials of the barycenter problem. Suppose we have $\mathcal{B}$-smooth\footnote{We say that a diffirentiable function $f:\mathbb{R}^{D}\rightarrow\mathbb{R}$ is $\mathcal{B}$-smooth if its gradient $\nabla f$ is $\mathcal{B}$-Lipschitz.} convex potentials $\{\psi_{n}\}$ for some $\mathcal{B}\in[0,+\infty]$, and denote
$\Delta=\text{\normalfont MultiCorr}(\{\alpha_{n}, \mathbb{P}_{n}\}\mid\{\psi_{n}\})-\text{\normalfont MultiCorr}(\{\alpha_{n}, \mathbb{P}_{n}\}\mid\{\psi^{*}_{n}\})$.
% \justin{where is this notion defined?} 
%  \lingxiao{should probably just define $\Delta$ to be the difference; I think $\Delta$ is a better notation than $\Delta$ which screams positivity}.
Then,
% \lingxiao{I changed the notations here quite a bit; update appendix}
\begin{eqnarray}
\Delta+\underbrace{\int_{\mathbb{R}^{D}}\sum_{n=1}^{N}\big[\alpha_{n}\overline{\psi_{n}}(y)-\frac{\|y\|^{2}}{2}\big]d\overline{\mathbb{P}}(y)}_{\text{Congruence mismatch}}\geq 
\frac{1}{2\mathcal{B}}\sum_{n=1}^{N}\alpha_{n}\|\nabla\psi_{n}^{*}(x)-\nabla\psi_{n}(x)\|^{2}_{\mathbb{P}_{n}}.
\label{cong-generative}
\end{eqnarray}
Here $\|\cdot\|_{\mu}$ denotes the norm induced by inner product in Hilbert space $\mathcal{L}^{2}(\mathbb{R}^{D}\rightarrow\mathbb{R}^{D},\mu)$. 
We call the second term on the left of \eqref{cong-generative} the \textbf{congruence mismatch}.
%\lingxiao{Wouldn't it make more sense to put congruence mismatch term on the left? After all you just want to bound the norm of the difference of gradient}
% \lingxiao{Be consistent with the space you are integrating over. What are $\mathbb{R}^{D}, \mathbb{R}^{D}$?}
\label{thm-main}
\end{theorem}
We prove this in Appendix \ref{sec-proofs}.
% \justin{can you give any intuition for what this theorem is saying?} - ALEX: explanation is below (I suppose you wrote this comment before reading it:)
%Theorem \ref{thm-main} explains several important aspects of multiple correlation.
% of optimization of correlations \lingxiao{this is vague: explain ``optimization of correlations''}. 
%First, if functions $\{\psi_{n}^{\dagger}\}$ are congruent, then the congruence mismatch vanishes and hence $\Delta\geq 0$. 
%This again confirms that the true potentials $\{\psi_{n}^{*}\}$ minimize multiple correlation within the class of congruent potentials. 
Note that if the congruence mismatch is non-positive, then 
\begin{equation}\Delta\geq \frac{1}{2\mathcal{B}}\sum_{n=1}^{N}\alpha_{n}\|\nabla\psi_{n}^{*}(x)-\nabla\psi_{n}(x)\|^{2}_{\mathbb{P}_{n}}\geq \frac{1}{\mathcal{B}}\sum_{n=1}^{N}\alpha_{n}\mathbb{W}_{2}^{2}(\nabla\psi_{n}\sharp \mathbb{P}_{n},\overline{\mathbb{P}}),
\label{cong-generative-reduced}
\end{equation}
where the last inequality of \eqref{cong-generative-reduced} follows from \cite[Lemma A.2]{korotin2019wasserstein}. From \eqref{cong-generative-reduced}, we conclude that for all $n\in\{1,\ldots,N\}$, we have 
$\mathbb{W}_{2}^{2}(\nabla\psi_{n}\sharp \mathbb{P}_{n},\overline{\mathbb{P}})\leq \frac{\mathcal{B}\Delta}{\alpha_{n}}.$ % We call this feature the \textbf{generative property};\lingxiao{I think it's best to avoid the term generative property. It doesn't match what you actually mean here, and there is no point in defining this term since you only refer to it like once or twice (you should just refer to the equation)} it provides a means of recovering the barycenter as the pushforward $\nabla\psi_{n}^{\dagger}\sharp \mathbb{P}_{n}$ after optimizing multiple correlation.% \justin{$\sharp$?}The gradient of the potential serves as the generator.
This shows that if the congruence mismatch is non-positive, then $\Delta$, the difference in multiple correlation, provides an upper bound for the Wasserstein-2 distance between the true barycenter and each pushforward $\nabla\psi_{n}\sharp \mathbb{P}_{n}$.
This justifies the use of $\nabla\psi_{n}\sharp \mathbb{P}_{n}$ to recover the barycenter. Notice for optimal potentials, the congruence mismatch is zero.

%From \eqref{cong-generative}, \eqref{cong-generative-reduced}, and discussion above, the generative property holds if the congruence mismatch is \textbf{negative}. 
Thus to penalize positive congruence mismatch, we introduce a regularizing term
\begin{equation}\mathcal{R}_{1}^{\overline{\mathbb{P}}}(\{\alpha_{n}\},\{\overline{\psi_{n}}\})\stackrel{\text{def}}{=}\int_{\mathbb{R}^{D}}\left[\sum_{n=1}^{N}\alpha_{n}\overline{\psi_{n}}(y)-\frac{\|y\|^{2}}{2}\right]_{+}d\overline{\mathbb{P}}(y).
\label{cheat-reg}
\end{equation}
Because we take the positive part of the integrand of \eqref{cong-generative} to get \eqref{cheat-reg} and that the right side of \eqref{cong-generative} is non-negative, we have
% The regularized objective upper-bounds the optimal congruence mismatch in \eqref{cong-generative}, 
% \lingxiao{I think it will be better to define congruence error more similar to (11), i.e., without $||\textup{id}||^2$}
$$\big[\text{\normalfont MultiCorr}(\{\alpha_{n},\mathbb{P}_{n}\}\mid\{\psi_{n}\})+1\cdot\mathcal{R}_{1}^{\overline{\mathbb{P}}}(\{\alpha_{n}\},\{\overline{\psi_{n}}\})\big]-\text{\normalfont MultiCorr}(\{\alpha_{n},\mathbb{P}_{n}\}\mid\{\psi^{*}_{n}\})\geq 0$$
% that the regularized multiple correlations necessarily upper bound the optimal multiple correlations \lingxiao{you should write this out; this implication is not so clear just by reading it especially you defined so many terms; should also include ``regularized MC for \textbf{any convex potentials} upperbound ...''}.
for all convex potentials $\{\psi_{n}\}$. 
On the other hand, for optimal potentials $\{\psi_{n}\}=\{\psi_{n}^{*}\}$, the inequality turns into equality, implying that adding the regularizing term $1\cdot\mathcal{R}_{1}^{\overline{\mathbb{P}}}(\{\alpha_{n}\},\{\overline{\psi_{n}}\})$ to \eqref{unreg-objective} will not introduce bias -- the optimal solution still yields $\{\psi_n^*\}$.

%\justin{this paragraph was hard for me to follow. try to give us the bottom line: what do these observations tell us about the barycenter you get?}

However, evaluating \eqref{cheat-reg} exactly requires knowing the true barycenter $\overline{\mathbb{P}}$ a priori. 
To remedy this issue, one may replace $\overline{\mathbb{P}}$ with another absolutely continuous measure $\tau\cdot\widehat{{\mathbb{P}}}$ ($\tau\geq 1$ and $\widehat{{\mathbb{P}}}$ is a probability measure) whose density bounds that of $\overline{\mathbb{P}}$ from above almost everywhere.
% \lingxiao{why is this always possible without making $\hat{\mathbb{P}}$'s support too large?}
In this case, %the regularizer $\mathcal{R}_{1}^{\overline{\mathbb{P}}}$ will upper bound the infeasible one, i.e.,
\begin{equation}\tau\cdot\mathcal{R}_{1}^{\widehat{{\mathbb{P}}}}(\{\alpha_{n}\},\{\overline{\psi_{n}}\})=\tau\cdot\int_{\mathbb{R}^{D}}\big[\sum_{n=1}^{N}\alpha_{n}\overline{\psi_{n}}(y)-\frac{\|y\|^{2}}{2}\big]_{+}d\widehat{{\mathbb{P}}}\geq \mathcal{R}_{1}^{\overline{\mathbb{P}}}(\{\alpha_{n}\},\{\overline{\psi_{n}}\}).
\label{ok-reg}
\end{equation}
%In this case, optimization of regularized multiple correlation satisfies identical properties to those discussed in the previous paragraph.\justin{say what properties} 
Hence we obtain the following regularized version of \eqref{unreg-objective} where $\{\psi_n^*\}$ is the optimal solution:
\begin{eqnarray}
\min_{\{\psi_{n}\}\in\text{Conv}}\big[\text{\normalfont MultiCorr}(\{\alpha_{n},\mathbb{P}_{n}\}\mid\{\psi_{n}\})+\tau\cdot\mathcal{R}_{1}^{\widehat{{\mathbb{P}}}}(\{\alpha_{n}\},\{\overline{\psi_{n}}\})\big].
\label{main-objective-unfinished}
\end{eqnarray}

Selecting a measure $\tau\cdot\widehat{{\mathbb{P}}}$ is not obvious. Consider the case when $\{\mathbb{P}_{n}\}$ are supported on compact sets $\mathcal{X}_{1},\ldots,\mathcal{X}_{N}\subset\mathbb{R}^{D}$ and $\mathbb{P}_{1}$ has density upper bounded by $h<\infty$. In this scenario, the barycenter density is upper bounded by $h\cdot \alpha_{1}^{-D}$ \cite[Remark 3.2]{alvarez2016fixed}. Thus, the measure $\tau\cdot\widehat{{\mathbb{P}}}$ supported on $\text{ConvexHull}(\mathcal{X}_{1},\dots,\mathcal{X}_{N})$ with this density is an upper bound for $\overline{\mathbb{P}}$. 
We will address the question of how to choose $\tau,\widehat{\mathbb{P}}$ properly in practice in \S\ref{sec-practical-aspects}.

%\justin{The last two or three paragraphs are likely to confuse the reader.  I think the issue is that the text is ``evolving'' an optimization problem by starting with multiple correlation and adding and removing terms.  It'd be helpful at the end to write an optimization problem, i.e., say something to the effect ``After introducing the simplifications above, we consider the following regularized problem for recovering the dual potentials...'' and actually state the minimization problem.  It's currently sort of implicit and doesn't appear until the next section, which makes the storytelling nonlinear.}

% Empirically, it turns out the method is not very sensible to the choice of $\nu$. Our experiments show that the optimization converges to the true barycenter even if $\nu$ does not upper bound $\overline{\mathbb{P}}$.

%\subsection{Algorithm and Theoretical Justification}
\subsection{Enforcing Conjugacy of Potentials Pairs}
Throughout this subsection, we assume the upper bound finite measure $\tau\cdot\widehat{{\mathbb{P}}}$ of the $\overline{\mathbb{P}}$ is known. 
The optimization problem \eqref{main-objective-unfinished}
% \lingxiao{the minimization notation below is weird; why not just put min in the front}
involves not only the potentials $\{\psi_{n}\}$, but also their conjugates $\{\overline{\psi_{n}}\}$. 
This brings practical difficulty since evaluating conjugate potentials is hard \citep{korotin2019wasserstein}.

Instead we parameterize potentials $\psi_{n}$ and $\overline{\psi_{n}}$ separately using input convex neural networks (ICNN) as $\psi_{n}^{\dagger}$ and $\overline{\psi_{n}^{\ddagger}}$ respectively. 
We add an additional cycle-consistency regularizer to enfore the conjugacy of $\psi_{n}^{\dagger}$ and $\overline{\psi_{n}^{\ddagger}}$ as in \citet{korotin2019wasserstein}.
% , because it has better scalability 
% \lingxiao{should mention briefly why it could scale better}.
% a pair of input convex neural networks (ICNNs) are used to approximate $\psi_{n}$ and $\overline{\psi_{n}}$. 
This regularizer is defined as
$$\mathcal{R}_{2}^{\mathbb{P}_{n}}(\psi_{n}^{\dagger}, \overline{\psi_{n}^{\ddagger}})\stackrel{\text{def}}{=}\int_{\mathbb{R}^{D}}\|\nabla\overline{\psi_{n}^{\ddagger}}\circ \nabla\psi_{n}^{\dagger}(x)-x\|_{2}^{2}\,d\mathbb{P}_{n}(x)=\|\nabla\overline{\psi_{n}^{\ddagger}}\circ \nabla\psi_{n}^{\dagger}-\text{id}_{\mathbb{R}^{D}}\|_{\mathbb{P}_{n}}^{2}.$$
Note that $\mathcal{R}_{2}^{\mathbb{P}_{n}}(\psi_{n}^{\dagger}, \overline{\psi_{n}^{\ddagger}}) = 0$ this condition is necessary for $\psi_{n}^{\dagger}$ and $ \overline{\psi_{n}^{\ddagger}}$ to be conjugate with each other. Also, it is a sufficient condition for convex functions to be conjugates up to an additive constant. 

We use one-sided regularization. In our case, computing the regularizer of the other direction $\|\nabla\psi_{n}^{\dagger}\circ \nabla\overline{\psi_{n}^{\ddagger}}-\text{id}_{\mathbb{R}^{D}}\|_{\overline{\mathbb{P}}}^{2}$ is infeasible, since $\overline{\mathbb{P}}$ is unknown. If fact, \citet{korotin2019wasserstein} demonstrates that such one-sided condition is sufficient. 

% \lingxiao{Please make sure my edits here are correct.}
%This regularizer checks that the gradients of the potentials are mutual inverses. This condition is necessary for two convex functions to be conjugate, and if this condition holds true, the functions are convex conjugates up to an additive constant. 
% constant \justin{shift? multiple?}.

In this way we use $2N$ input convex neural networks for $\{\psi^{\dagger}_{n}, \overline{\psi^{\ddagger}_{n}}\}$.
By adding the new cycle consistency regularizer into \eqref{main-objective-unfinished}, we obtain our final objective:
% \justin{give eqn number}.
% We use $2N$ input-convex neural networks for $\{\psi^{\dagger}_{n}, \overline{\psi^{\ddagger}_{n}}\}$, 
% \justin{are these missing subscripts?}
% resulting in the optimization problem
\begin{eqnarray}
\min_{\{\psi^{\dagger}_{n},\overline{\psi_{n}^{\ddagger}}\}}\!\!
\overbrace{
\sum_{n=1}^{N}\!\bigg[\!\alpha_{n}\!\!
\int_{\!\!\raisebox{-.05in}{\scriptsize$\mathbb{R}^{\!D}$}}\hspace{-.08in} % blame justin for these latex hacks
[
\underbrace{
\langle x,\nabla\psi_{n}^{\dagger}(x)\rangle\!-\!\overline{\psi_{n}^{\ddagger}}(\nabla\psi_{n}^{\dagger}(x))}_{\approx \psi_{n}^{\ddagger}(x)}
]
d\mathbb{P}_{n}(x)\!
\bigg]}^{\text{Approximate multiple correlation}}
\!\!+\!
%
%\nonumber
%\\
\underbrace{\tau\!\cdot\!\mathcal{R}_{1}^{\widehat{{\mathbb{P}}}}(\{\overline{\psi_{n}^{\ddagger}}\})}_{\text{Congruence reg.}}
\!+\!
\underbrace{\lambda\!\!\sum_{n=1}^{N}\!\alpha_{n} \mathcal{R}_{2}^{\mathbb{P}_{n}}(\psi_{n}^{\dagger}, \overline{\psi_{n}^{\ddagger}})}_{\text{Cycle regularizer}}\!.
\label{main-objective}
\end{eqnarray}
Note that we express the aproximate multiple correlation by using both potentials $\{\psi^{\dagger}_{n}\}$ and $\{\overline{\psi^{\ddagger}_{n}}\}$. This is done to eliminate the freedom of an additive constant on $\{\psi_n^\dagger\}$ that is not addressed by cycle regularization.
% \lingxiao{You should explain why you approximate MultiCorr with the first term in \eqref{main-objective} to remove the freedom of an additive constant on $\psi_n^\dagger$.}
We denote the entire objective as $\text{MultiCorr}\big(\{\mathbb{P}_{n}\}\mbox{ }\vert\mbox{ } \{\psi^{\dagger}\}, \{\overline{\psi^{\ddagger}}\}; \tau,\widehat{{\mathbb{P}}}, \lambda\big)$. 
Analogous to Theorem \ref{thm-main}, we have following result showing that this new objective enjoys the same properties as the unregularized version from \eqref{unreg-objective}.  
\begin{theorem}%[Generative Property for Approximators of Multiple Correlations]
Let $\overline{\mathbb{P}}\in\mathcal{P}_{2,ac}(\mathbb{R}^{D})$ be the barycenter of $\mathbb{P}_{1},\dots,\mathbb{P}_{N}\in\mathcal{P}_{2,ac}(\mathbb{R}^{D})$ w.r.t.\ weights $\alpha_{1},\dots,\alpha_{N}$. Let $\{\psi_{n}^{*}\}$ be the optimal congruent potentials of the barycenter problem. Suppose we have $\tau, \hat{\mathbb{P}}$ such that $\tau \geq 1$ and $\tau\cdot\widehat{{\mathbb{P}}}\geq\overline{\mathbb{P}}$. %Moreover, assume that convex potentials $\{\psi_{n}^{\dagger},\overline{\psi_{n}^{\ddagger}}\}$ satisfy
%\label{eps-def-thm2}
%\end{equation}
%for some $\Delta\in\mathbb{R}$. 
Suppose we have convex potentials $\{\psi_n^\dagger\}$ and $\beta^{\ddagger}$-strongly convex and $\mathcal{B}^{\ddagger}$-smooth convex potentials $\{\overline{\psi_{n}^{\ddagger}}\}$ with  ${0<\beta^{\ddagger}\leq\mathcal{B}^{\ddagger}<\infty}$ and $\lambda>\frac{\mathcal{B}^{\dagger}}{2(\beta^{\ddagger})^{2}}$. 
Then%, we can bound multiple correlation as follows:
%\begin{enumerate}[wide, labelwidth=0pt, labelindent=0pt]
%\item \textbf{Multiple Correlations Upper Bound} (reg. correlations dominate the true ones, i.e. $\Delta\geq 0$):
\begin{equation}\text{MultiCorr}\big(\{\alpha_{n},\mathbb{P}_{n}\}\mbox{ }\vert\mbox{ } \{\psi^{\dagger}_{n}\}, \{\overline{\psi^{\ddagger}_{n}}\}; \tau, \widehat{{\mathbb{P}}}, \lambda\big)\geq \text{MultiCorr}\big(\{\alpha_{n},\mathbb{P}_{n}\}\mbox{ }\vert\mbox{ } \{\psi^{*}_{n}\} \big).
\label{multicorr-upper-bound}
\end{equation}
%\item \textbf{Generative Property of Potentials} (for all $n=1,\dots,N$)
Denote
$\Delta=\text{MultiCorr}\big(\{\alpha_{n},\mathbb{P}_{n}\}\mbox{ }\vert\mbox{ } \{\psi^{\dagger}_{n}\}, \{\overline{\psi^{\ddagger}_{n}}\}; \tau, \widehat{{\mathbb{P}}}, \lambda\big)-\text{MultiCorr}\big(\{\alpha_{n},\mathbb{P}_{n}\}\mbox{ }\vert\mbox{ } \{\psi^{*}_{n}\}\big).$ 
Then for all $n\in\{1,\ldots,N\}$, we have
\begin{eqnarray}\mathbb{W}_{2}^{2}\big(\nabla\psi_{n}^{\dagger}\sharp\mathbb{P}_{n},\overline{\mathbb{P}}\big)\leq  \frac{2\Delta}{\alpha_{n}}\cdot\left(\sqrt{\frac{1}{\beta^{\ddagger}}}+\sqrt{\frac{1}{\lambda(\beta^{\ddagger})^{2}-\frac{\mathcal{B}^{\dagger}}{2}}}\right)^{2}=O(\Delta).
\label{cong-generative-regularized}
\end{eqnarray}
%\end{enumerate}
\label{thm-main-top}
\end{theorem}
% \justin{oops, i ended up using $\beta$ above, so I replaced $\beta^\ddagger$ with $\gamma^\ddagger$. sorry.}
Informally, Theorem \ref{thm-main-top} states that the better we solve the regularized dual problem, \eqref{main-objective}
% \justin{equation number?} 
%for $\{\mathbb{P}_{n}\}$ by potentials $\{\psi^{\dagger}_{n},\overline{\psi_{n}^{\ddagger}}\}$, 
the closer we expect each $\nabla\psi_{n}^{\dagger}\sharp \mathbb{P}_{n}$ 
% \lingxiao{should be $\nabla \psi_n^\dagger$}
to be to the true barycenter $\overline{\mathbb{P}}$ in $\mathbb{W}_{2}$. 
It follows from \eqref{multicorr-upper-bound} that our final objective \eqref{main-objective} is \emph{unbiased}: the optimal solution is obtained by $\{\psi^{*}_{n},\overline{\psi_{n}^{*}}\}$.
%The multiple regularized correlation gives a \emph{tight} upper bound: if we substitute the optimal $\{\psi^{\dagger}_{n},\overline{\psi_{n}^{\ddagger}}\}=\{\psi^{*}_{n},\overline{\psi_{n}^{*}}\}$, we get $\Delta=0$.
% \lingxiao{I'm not sure what the last sentence means. Do you just mean \ref{main-objective} is unbiased? Consider removing} ALEX: \beta -> \Delta

% Our numerical optimization procedure for computing the barycenter of measures $\{\mathbb{P}_{n}\}$ given by implicit samplers is provided in Appendix \ref{sec-algorithm-procedure}. 
%We parametrize potentials $\{\psi^{\dagger}_{n},\psi^{\ddagger}_{n}\}$ by input convex neural networks. The optimization \eqref{main-objective} is done via mini-batch stochastic gradient descent by using Monte-Carlo estimates by random samples from input measures $\{\mathbb{P}_{n}\}$ and regularization measure $\hat{\mathbb{P}}$.

\subsection{Practical Aspects and Optimization Procedure}
\label{sec-practical-aspects}
% For convenience, we decompose $\nu=\tau\cdot\nu_{0}$, where $\tau>0$ and $\nu_{0}$ is a probability measure. As we noted earlier, the only non-trivial question in the optimization multiple correlations is the choice of $\nu$. In general case, such a choice seems infeasible.

% \justin{say a bit more.  is it SGD?  what are the inputs and outputs?  help the readers who aren't theoretically oriented}
In practice, even if the choice of $\tau, \widehat{{\mathbb{P}}}$ does not satisfy $\tau\cdot\widehat{{\mathbb{P}}}\geq\overline{\mathbb{P}}$, we observe the pushforward measures $\nabla\psi_{n}^{\dagger}\sharp \mathbb{P}_{n}$ often converge to $\overline{\mathbb{P}}$.
To partially bridge the gap between theory and practice, we \emph{dynamically} update the measure $\widehat{\mathbb{P}}$ so that after each optimization step we set (for $\gamma\in[0,1]$)%\footnote{
%As the initial measure $\widehat{\mathbb{P}}$ one may use the barycenter of $\{\mathcal{N}(\mu_{\mathbb{P}_{n}}, \Sigma_{\mathbb{P}_{n}})\}$. It can be efficiently computed via iterative fixed point algorithm \citep{alvarez2016fixed,chewi2020gradient}.}
$$\widehat{\mathbb{P}}':=\gamma\cdot\widehat{\mathbb{P}}+(1-\gamma)\cdot\sum_{n=1}^{N}\alpha_{n}\cdot \big[\nabla\psi^{\dagger}\sharp\mathbb{P}_{n}\big],$$
i.e., the probability measure $\widehat{\mathbb{P}}'$ is a mixture of the given initial measure $\widehat{\mathbb{P}}$ and the current barycenter estimates $\{\nabla\psi^{\dagger}\sharp\mathbb{P}_{n}\}$. 
For the initial $\widehat{\mathbb{P}}$ one may use the barycenter of $\{\mathcal{N}(\mu_{\mathbb{P}_{n}}, \Sigma_{\mathbb{P}_{n}})\}$. It can be efficiently computed via an iterative fixed point algorithm \citep{alvarez2016fixed,chewi2020gradient}.
During the optimization, these estimates become closer to the true barycenter and can thus improve the congruence regularizer \eqref{ok-reg}. 
% \lingxiao{The bolding here seems unnecessary.}

We use mini-batch stochastic gradient descent to solve \eqref{main-objective} where the integration is done by Monte-Carlo sampling from input measures $\{\mathbb{P}_{n}\}$ and regularization measure $\widehat{\mathbb{P}}$, similar to \citet{li2020continuous}.
We provide the detailed optimization procedure (Algorithm \ref{algorithm-main}) and discuss its computational complexity in Appendix \ref{sec-algorithm-procedure}. In Appendix \ref{sec-cycle-cong-check}, we demonstrate that the impact of the considered regularization on our model: we show that cycle consistency and the congruence condition of the potentials are well satisfied.

% In ageneral, we do not know how to choose such a measure that will necessarily upper bound the barycenter measure. However, we emphasize that in some cases, it is possible to provide

% Let $\mathbb{P}_{1},\dots,\mathbb{P}_{N}$ be continuous probability distributions with finite second moment on $\mathcal{X}\subset \mathbb{R}^{D}$. Let $\bm{\alpha}=(\alpha_{1},\dots,\alpha_{N})$ be a weight vector, i.e. $\bm{\alpha}>0$, $\|\bm{\alpha}\|_{1}=1$. We denote the Wasserstein-2 barycenter of $\mathbb{P}_{1},\dots,\mathbb{P}_{N}$ w.r.t. weights $\alpha_{1},\dots,\alpha_{n}$ by $\overline{\mathbb{P}}$, i.e.
% $$\overline{\mathbb{P}}=\argmin_{\mathbb{P}}\sum_{n=1}^{N}\alpha_{n}\mathbb{W}_{2}^{2}(\mathbb{P}_{n},\mathbb{P}).$$

% Assume that the optimal coupling between $\mathbb{P}$ and each $\mathbb{P}_{n}$ is deterministic (it is possible to impose specific conditions when it is necessarily deterministic). Thus, if has a form of a gradient of a convex function: $\nabla\overline{\psi^{*}}\circ\overline{\mathbb{P}}=\mathbb{P}_{n}$. We also consider the inverse mappings $\nabla\psi^{*}\circ\mathbb{P}_{n}=\overline{\mathbb{P}}$. Recall that the key property of the Wasserstein-2 barycenter is that
% $$\sum_{n=1}^{N}\alpha_{n}\nabla\overline{\psi^{*}}(x)=x,$$
% that is also equivalent to 
% $$\sum_{n=1}^{N}\alpha_{n}\overline{\psi^{*}}(x)=\frac{x^{2}}{2}+c.$$
\section{Experiments}
\label{sec-experiments}
The code is written on \textbf{PyTorch} framework and is publicly available at
\begin{center}
\url{https://github.com/iamalexkorotin/Wasserstein2Barycenters}.
\end{center}

We compare our method [C$\mathbb{W}_{2}$B] with the potential-based method [CR$\mathbb{W}$B] by \citet{li2020continuous} (with Wasserstein-2 distance and $\mathcal{L}^{2}$-regularization) and with the measure-based generative method [SC$\mathbb{W}_2$B] by \citet{fan2020scalable}.
%  The latter is a case of the generic algorithm by \citep{cohen2020estimating} with the optimal transport solver by \citep{makkuva2019optimal}.
 All considered methods recover $2N$ potentials ${\{\psi^{\dagger}_{n},\overline{\psi_{n}^{\ddagger}}\}\approx \{\psi_{n}^{*},\overline{\psi_{n}^{*}}\}}$ and approximate the barycenter as pushforward measures $\{\nabla\psi^{\dagger}_{n}\sharp\mathbb{P}_{n}\}$. 
 Regularization in [CR$\mathbb{W}$B] allows access to the joint density of the transport plan, a feature of their method that we do not consider here.
 The method [SC$\mathbb{W}_2$B] additionally outputs a generated barycenter $g\sharp \mathbb{S}\approx \overline{\mathbb{P}}$ where $g$ is the generative network and $\mathbb{S}$ is the input noise distribution.
%  \justin{replace $\mathbb Z$ with something else}

 To assess the quality of the computed barycenter, we consider the \textbf{unexplained variance percentage} defined as 
%  \lingxiao{Perhaps explain the choice of name unexplained variance percentage a bit more here}
$\mbox{UVP}(\tilde{\mathbb{P}})=100\frac{\mathbb{W}_{2}^{2}(\tilde{\mathbb{P}},\overline{\mathbb{P}})}{\sfrac{1}{2}\mbox{\small Var}(\overline{\mathbb{P}})}\%.$
When $\mbox{UVP}\approx 0\%$, $\tilde{\mathbb{P}}$ is a good approximation of $\overline{\mathbb{P}}$. 
For values $\geq 100\%$, the distribution $\tilde{\mathbb{P}}$ is undesirable: a trivial baseline $\mathbb{P}^{0}=\delta_{\mathbb{E}_{\overline{\mathbb{P}}}[y]}$ achieves $\mbox{UVP}(\mathbb{P}^{0})=100\%$.
 Evaluating UVP in high dimensions is infeasible: empirical estimates of $\mathbb{W}_{2}^{2}$ are unreliable due to high sample complexity %.\footnote{The sample complexity of $\mathbb{W}_{2}^{2}$ is $O(K^{-\frac{1}{D}})$ where $K$ is the number of samples, see 
\citep{weed2019sharp}. % }
% highly biased \lingxiao{explain which term is biased? the W2 term?}.
To overcome this issue, for barycenters given by $\nabla\psi^{\dagger}_{n}\sharp\mathbb{P}_{n}$ we use $\mathcal{L}^{2}\mbox{-UVP}$ defined by

\vspace{-3mm}\begin{equation}
\mathcal{L}^{2}\mbox{-UVP}(\nabla\psi^{\dagger}_{n},\mathbb{P}_{n})\stackrel{\text{def}}{=}100\frac{\|\nabla\psi^{\dagger}_{n}-\nabla\psi^{*}_{n}\|_{\mathbb{P}_{n}}^{2}}{\mbox{\small Var}(\overline{\mathbb{P}})}\%\qquad\bigg[\geq \mbox{UVP}(\nabla \psi_n^\dagger \sharp \mathbb{P}_n)\bigg],
\label{l2-uvp-def}
\end{equation}
%\vspace{-3mm}
where the inequality in brackets follows from \citep[Lemma A.2]{korotin2019wasserstein}.
%It compares transport maps $\nabla\psi^{\dagger}_{n}$ and $\nabla\psi^{*}_{n}$ as functions by using $\mathcal{L}^{2}(\mathbb{R}^{D}\rightarrow\mathbb{R}^{D},\mathbb{P}_{n})$ norm. $\mathcal{L}^{2}\mbox{-UVP}$ upper-bounds UVP. 
%\lingxiao{TODO: cite and explain the upper bound inequality, and fix appendix}
We report the weighted average of $\mathcal{L}^{2}\mbox{-UVP}$ of all pushforward measures w.r.t.\ the weights $\alpha_{n}$. 
For barycenters given in an implicit form $g\sharp\mathbb{S}$, 
% \justin{$\sharp$?}
we compute the \textbf{Bures-Wasserstein} UVP defined by
%\vspace{-3mm}
\begin{equation}\text{B}\mathbb{W}_{2}^{2}\text{-UVP}(g\sharp\mathbb{S})\stackrel{\text{def}}{=}100\frac{\text{B}\mathbb{W}_{2}^{2}(g\sharp\mathbb{S},\overline{\mathbb{P}})}{\frac{1}{2}\mbox{\small Var}(\overline{\mathbb{P}})}\%\qquad\bigg[\leq \mbox{UVP}(g\sharp \mathbb{S})\bigg],
\label{bw-uvp-def}
\end{equation}
where $\text{B}\mathbb{W}_{2}^{2}(\mathbb{P},\mathbb{Q})=\mathbb{W}_{2}^{2}\big(\mathcal{N}(\mu_{\mathbb{P}},\Sigma_{\mathbb{P}}),\mathcal{N}(\mu_{\mathbb{Q}},\Sigma_{\mathbb{Q}})\big)$ is the Bures-Wasserstein metric and we use $\mu_\mathbb{P}, \Sigma_{\mathbb{P}}$ to denote the mean and the covariance of a distribution $\mathbb{P}$ \citep{chewi2020gradient}.
It is known that $\text{B}\mathbb{W}_{2}^{2}$ lower-bounds $\mathbb{W}_2^2$ \citep{dowson1982frechet}, so the inequality in the brackets of \eqref{bw-uvp-def} follows.
A detailed discussion of the adopted metrics is given in Appendix \ref{sec-metrics}.
% \lingxiao{briefly comment on what $\text{B}\mathbb{W}_{2}^{2}\text{-UVP}$ is, and why ther are upper/lower bound}
%We discuss $\mathcal{L}^{2}\mbox{-UVP}$ and $\text{B}\mathbb{W}_{2}^{2}\text{-UVP}$ in more detail and provide their closed forms in Appendix \ref{sec-metrics}: see \eqref{l2-uvp-def} and \eqref{bw-uvp-def}.

% Justin removed the paragraph below, didn't add anything
% In \S\ref{sec-exp-hd}, we test the method in high-dimensional location-scatter cases, where a ground truth barycenter is available.
% In \S\ref{sec-subset-posterior}, we test subset posterior aggregation \citep{srivastava2015wasp}.

% \lingxiao{should insert some short sentences here about toy examples like swiss-roll}

\subsection{High-Dimensional Location-Scatter Experiments}
\label{sec-exp-hd}

\begin{figure}[!h]
     \centering
     \begin{subfigure}[b]{0.28\columnwidth}
         \centering
         \includegraphics[width=\textwidth]{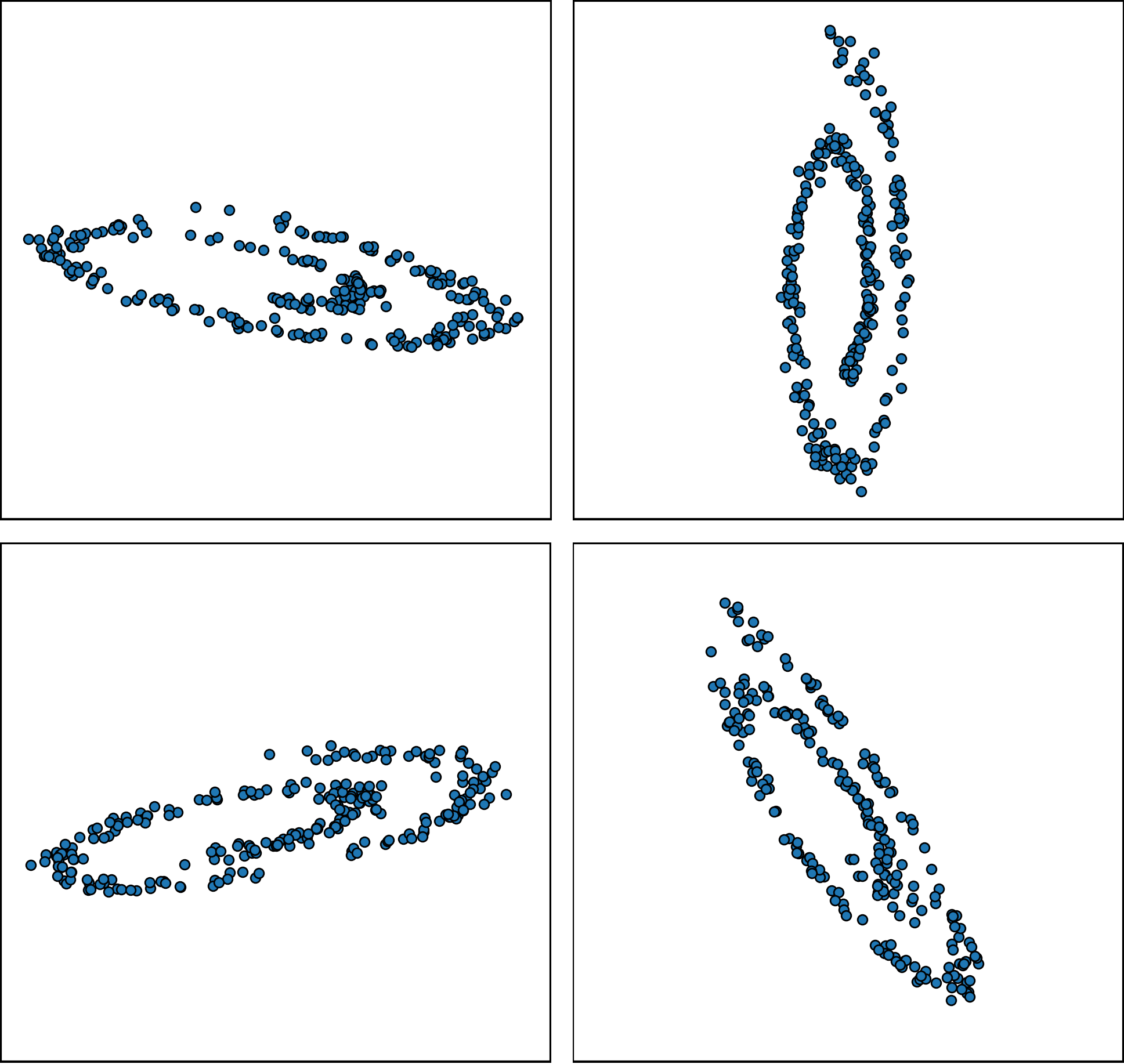}
         \caption{Input distributions $\{\mathbb{P}_{n}\}$}
        \label{fig:ls-sr-dim2-inputs}
     \end{subfigure}
     \hfill
     \begin{subfigure}[b]{0.28\columnwidth}
         \centering
         \includegraphics[width=\linewidth]{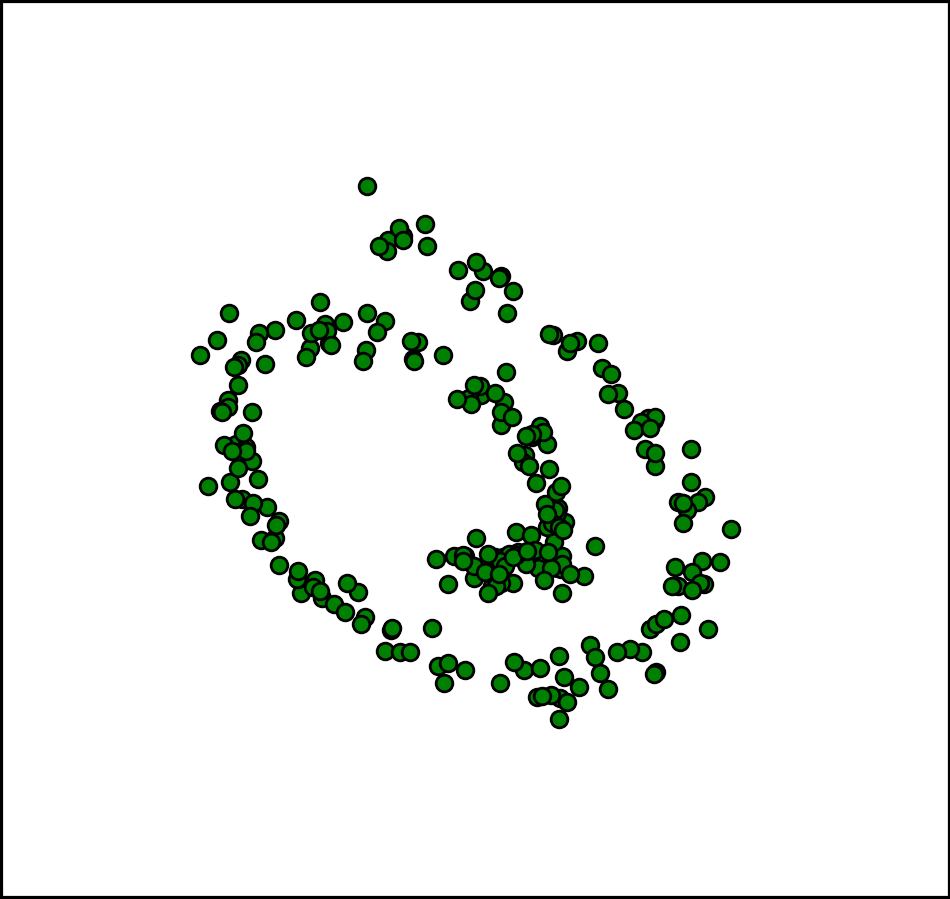}
     \caption{True barycenter $\overline{\mathbb{P}}$}
        \label{fig:ls-sr-dim2-bar}
     \end{subfigure}
     \hfill
     \begin{subfigure}[b]{0.28\columnwidth}
         \centering
         \includegraphics[width=\linewidth]{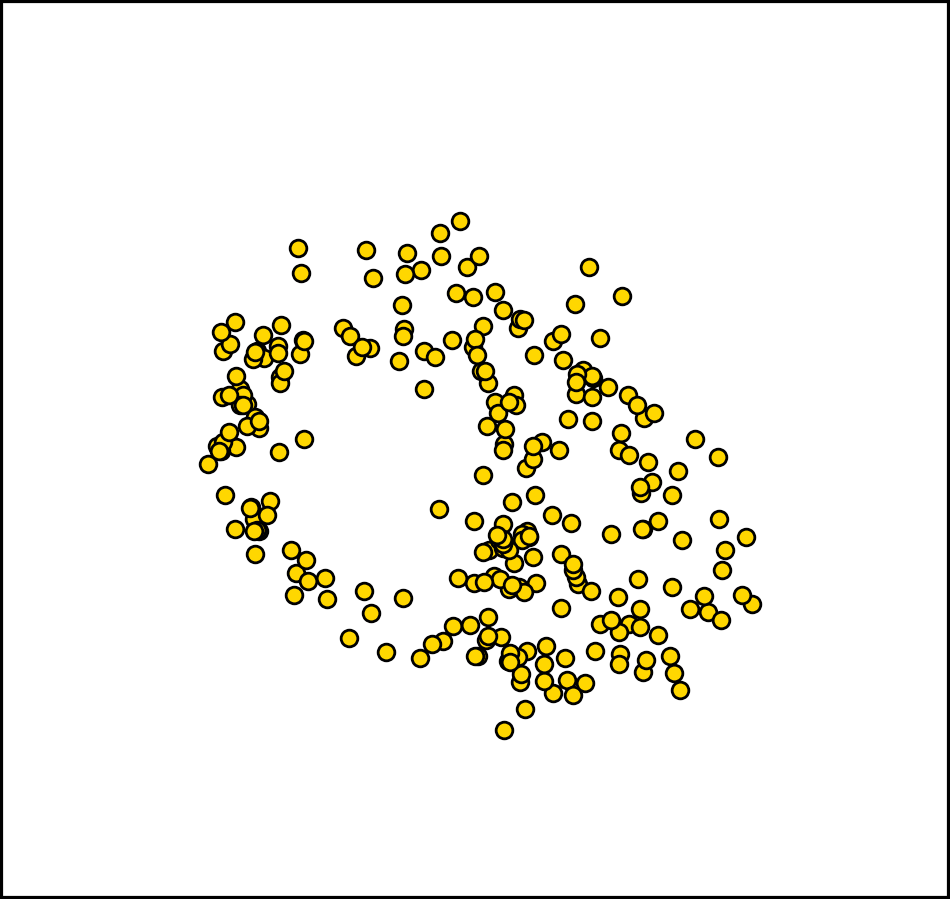}
         \caption{SC$\mathbb{W}_2$B, generated distribution $g\sharp\mathbb{S}$}
        \label{fig:ls-sr-dim2-gen}
     \end{subfigure}
     
     \vspace{2mm}
     \begin{subfigure}[b]{0.28\columnwidth}
         \centering
         \includegraphics[width=\linewidth]{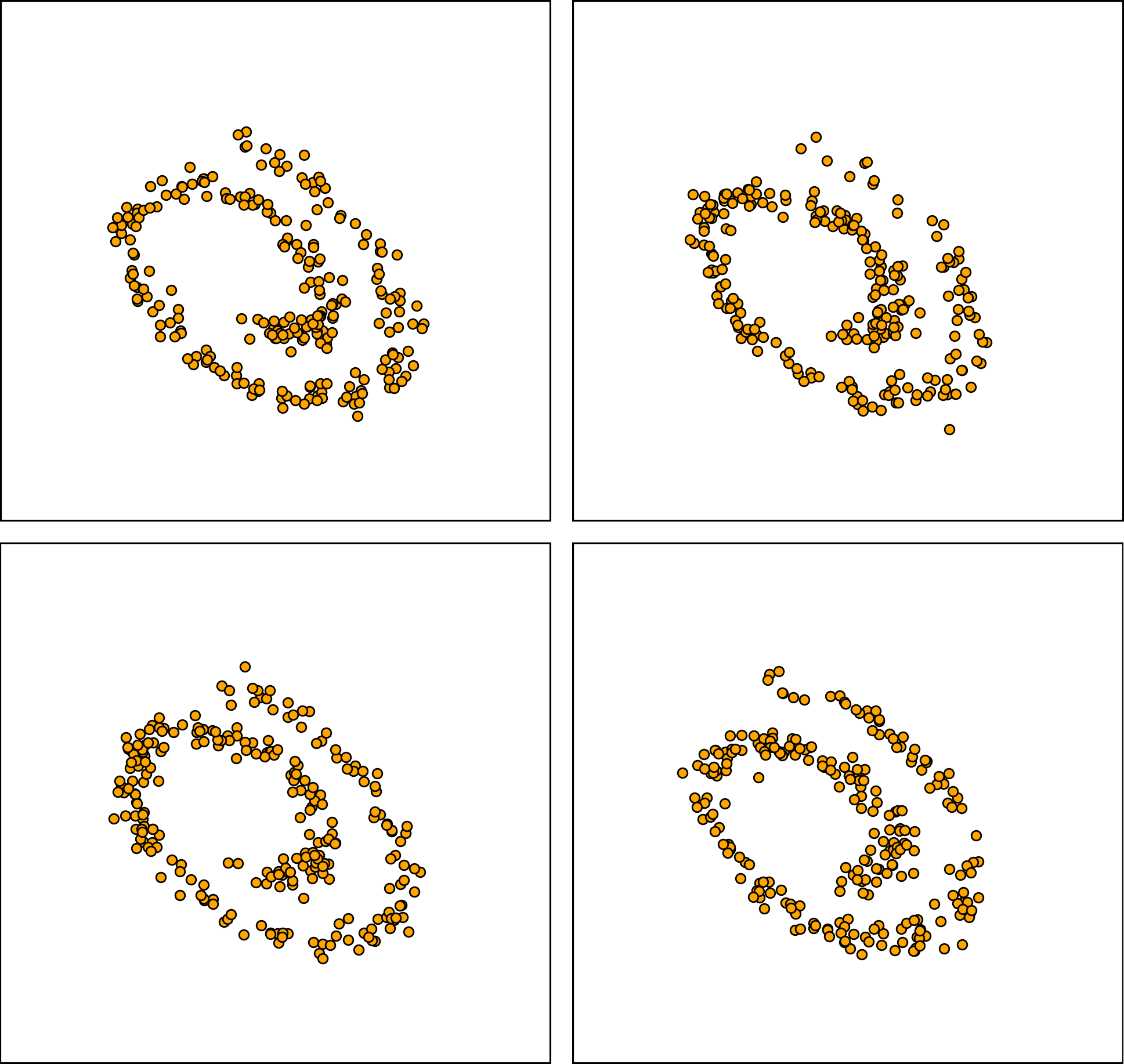}
         \caption{SC$\mathbb{W}_2$B, distributions $\nabla\psi_{n}^{\dagger}\sharp\mathbb{P}_{n}$}
        \label{fig:ls-sr-dim2-scwb}
     \end{subfigure}
     \hfill
     \begin{subfigure}[b]{0.28\columnwidth}
         \centering
         \includegraphics[width=\linewidth]{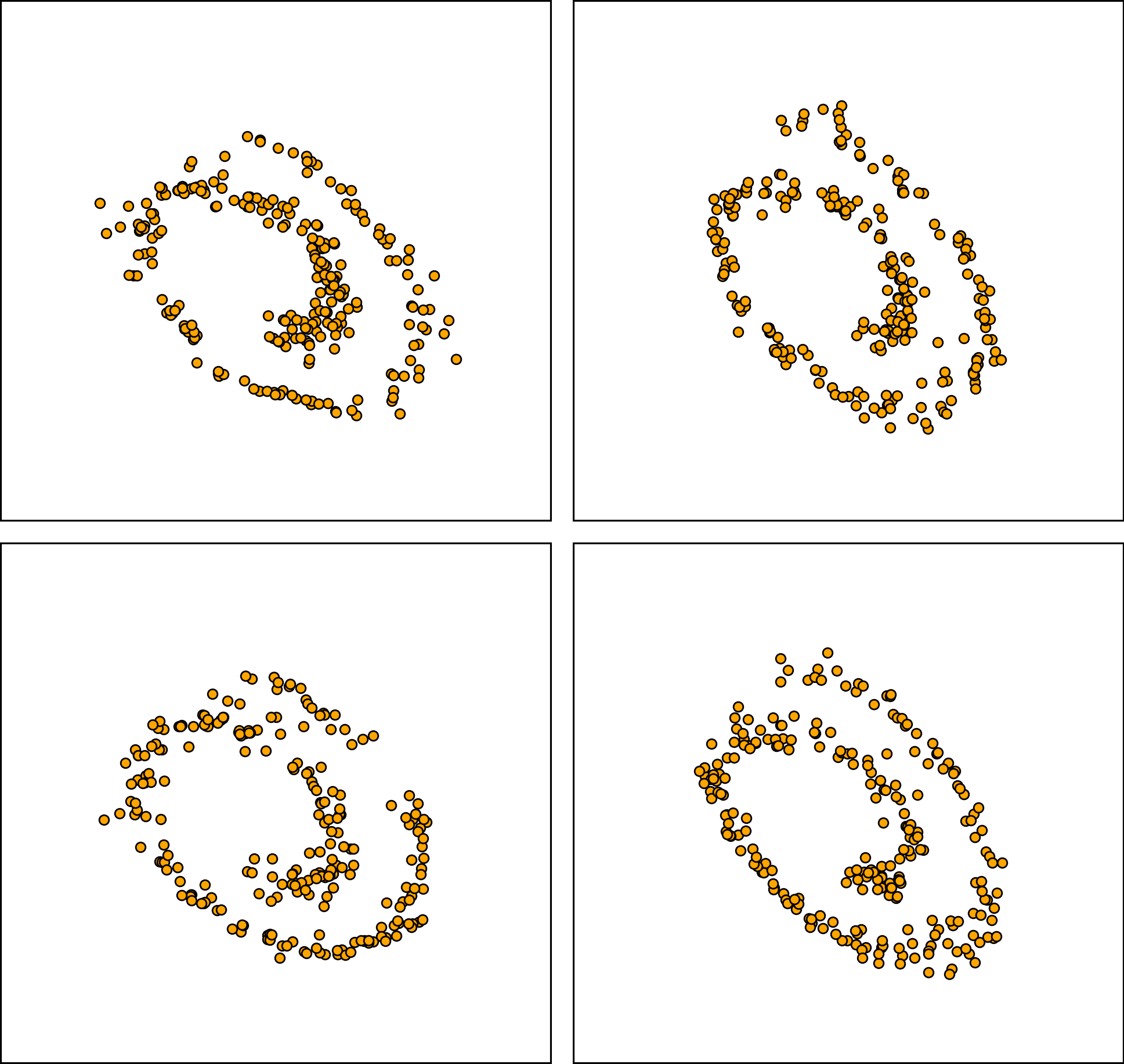}
         \caption{CR$\mathbb{W}$B, distributions ${\nabla\psi_{n}^{\dagger}\sharp\mathbb{P}_{n}}$}
        \label{fig:ls-sr-dim2-crwb}
     \end{subfigure}
     \hfill
     \begin{subfigure}[b]{0.28\columnwidth}
         \centering
         \includegraphics[width=\linewidth]{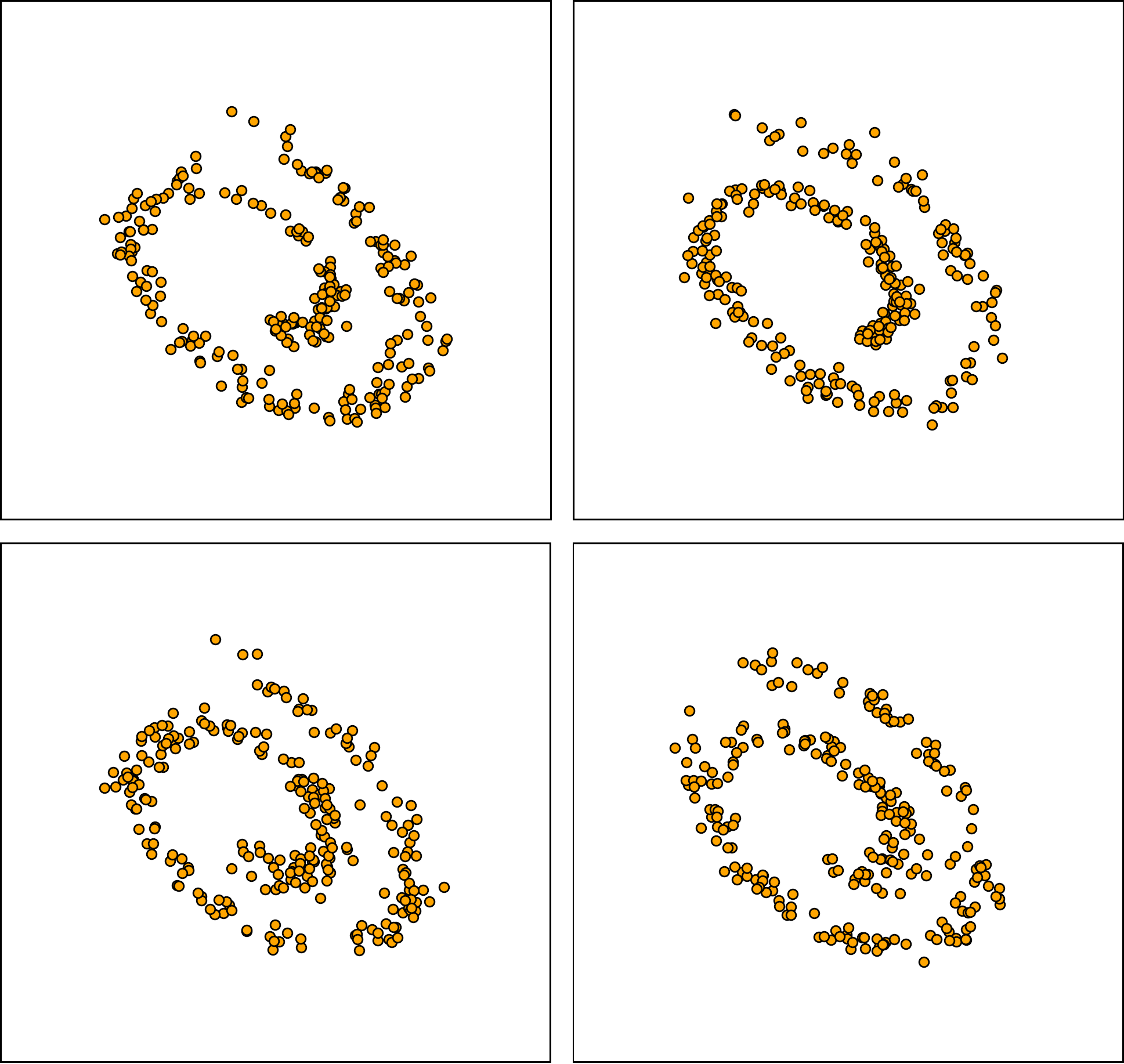}
         \caption{C$\mathbb{W}_{2}$B, distributions ${\nabla\psi_{n}^{\dagger}\sharp\mathbb{P}_{n}}$}
        \label{fig:ls-sr-dim2-w2cb}
     \end{subfigure}
    %  \vspace{-.1in}
    \caption{Barycenter of location-scatter Swiss roll population computed by three methods.}
    \vspace{-.1in}
    % \justin{can save space by eliminating white space around the swiss rolls}}
    \label{fig:ls-sr-dim2}
\end{figure}

In this section, we consider $N=4$ with $(\alpha_{1},\dots,\alpha_{4})=(0.1, 0.2, 0.3, 0.4)$ as weights.
We consider the \textbf{location-scatter family} of distributions \cite[\S4]{alvarez2016fixed} whose true barycenter can be computed. Let $\mathbb{P}_{0}\in\mathcal{P}_{2,\text{ac}}$ and define the following location-scatter family of distributions
$\mathcal{F}(\mathbb{P}_{0})=\{f_{S,u}\sharp \mathbb{P}_{0}\mbox{ }\vert\mbox{ } S\in\mathcal{M}^{+}_{D\times D}, u\in\mathbb{R}^{D}\},$
% \footnote{If $\mathbb{P}_{0}$ is a non-degenerate Gaussian distribution, then $\mathcal{F}(\mathbb{P}_{0})$ is the set of all non-degenerate Gaussians.}
where $f_{S,u}:\mathbb{R}^{D}\rightarrow\mathbb{R}^{D}$ is a linear map $f_{S,u}(x)=Sx+u$ with positive definite matrix $S\in\mathcal{M}^{+}_{D\times D}$.
% For each $\Sigma\in \mathcal{M}^{+}_{D\times D}$ there exists a unique $\mathbb{P}\in \mathcal{F}(\mathbb{P}_{0})$ having $\Sigma$ as its covariance matrix. Moreover, for every $\mathbb{P}\in \mathcal{F}(\mathbb{P}_{0})$ there exists a unique pair $(S,u)\in \mathcal{M}^{+}_{D\times D}\times\mathbb{R}^{D}$ such that $f_{S,u}\circ\mathbb{P}_{0}=\mathbb{P}$.
When $\{\mathbb{P}_{n}\}\subset\mathcal{F}(\mathbb{P}_{0})$, their barycenter $\overline{\mathbb{P}}$ is also an element of $\mathcal{F}(\mathbb{P}_{0})$ and can be computed via fixed-point iterations \citep{alvarez2016fixed}.

Figure \ref{fig:ls-sr-dim2-inputs} shows a 2-dimensional location-scatter family generated by using the Swiss roll distribution as $\mathbb{P}_0$. 
% \justin{I couldn't parse the following sentence:}
The true barycenter is shown in Figure \ref{fig:ls-sr-dim2-bar}. 
The generated barycenter $g\sharp \mathbb{S}$ of [SC$\mathbb{W}_2$B] is given in Figure \ref{fig:ls-sr-dim2-gen}. 
The pushforward measures $\nabla \psi_n^\dagger\sharp\mathbb{P}_n$ of each method are provided in Figures \ref{fig:ls-sr-dim2-scwb}, \ref{fig:ls-sr-dim2-crwb}, \ref{fig:ls-sr-dim2-w2cb}, respectively. 
In this example, the pushforward measures $\nabla\psi_{n}\sharp\mathbb{P}_{n}$ all reasonably approximate $\overline{\mathbb{P}}$, whereas the generated barycenter $g\sharp\mathbb{S}$ of [SC$\mathbb{W}_2$B] (Figure \ref{fig:ls-sr-dim2-gen}) visibly underfits.

For quantitative comparison, we consider two choices for $\mathbb{P}_{0}$: the $D$-dimensional standard Gaussian distribution and the uniform distribution on $[-\sqrt{3},+\sqrt{3}]^{D}$. 
Each $\mathbb{P}_{n}$ is constructed as $f_{S_{n}^{T}\Lambda S_{n},0}\sharp\mathbb{P}_{0}\in\mathcal{F}(\mathbb{P}_{0})$, where $S_{n}$ is a random rotation matrix and $\Lambda$ is diagonal with entries $[\frac{1}{2}b^0,\frac{1}{2}b^{1},\dots,2]$ where $b=\sqrt[D-1]{4}$.
We consider only centered distributions (i.e. zero mean) because the barycenter of non-centered ${\{\mathbb{P}_{n}\}\in\mathcal{P}_{2,\text{ac}}(\mathbb{R}^{D})}$ is the barycenter of $\{\mathbb{P}_{n}'\}$ shifted by $\sum_{n=1}^{N}\alpha_{n}\mu_{\mathbb{P}_{n}}$, where $\{\mathbb{P}_{n}'\}$ are centered copies of $\{\mathbb{P}_{n}\}$ \citep{alvarez2016fixed}. 
Results are shown in Table \ref{table-gaussians} and \ref{table-rectangles}.

% In Figure \ref{fig:ls-dim2-barycenters} we depict the barycenter of 2D location-scatter population computed by the considered methods.
\begin{table}[!ht]
\scriptsize
\centering
\hspace{-5mm}\begin{tabular}{|c|c|c|c|c|c|c|c|c|c|}
\hline
\textbf{Metric}                  & \textbf{Method}                & \textbf{D=2}    & \textbf{4}    & \textbf{8}    & \textbf{16}    & \textbf{32}    & \textbf{64}          & \textbf{128}         & \textbf{256}        \\ \hline
\multirow{2}{*}{B$\mathbb{W}_{2}^{2}$-UVP, \%}                  & [FC$\mathbb{W}$B], \cite{cuturi2014fast}                & 0.7   & 0.68    & 1.41   & 3.87    & 8.85    & 14.08          & 18.11         & \color{orange}{21.33}        \\ \cline{2-10}
  & \multirow{2}{*}{[SC$\mathbb{W}_2$B], \citep{fan2020scalable}} & 0.07 & 0.09 & 0.16 & 0.28  & 0.43  & 0.59        & 1.28        & 2.85       \\ \cline{1-1} \cline{3-10} 
\multirow{3}{*}{\shortstack{$\mathcal{L}_{2}$-UVP, \%\\(potentials)}} &                       & 0.08 & 0.10 & 0.17 & 0.29  & 0.47  & 0.63        & 1.14        & 1.50       \\ \cline{2-10} 
& [CR$\mathbb{W}$B], \citep{li2020continuous} & 0.99 & 2.52 & 8.62 & \color{orange}{22.23} & \color{red}{67.01} & \multicolumn{3}{c|}{\color{red}{>100}} \\ \cline{2-10} & [C$\mathbb{W}_{2}$B], \underline{\textbf{ours}}                & \color{green}{0.06} & \color{green}{0.05} & \color{green}{0.07} & \color{green}{0.11}  & \color{green}{0.19}  & \color{green}{0.24}        & \color{green}{0.42}        & \color{green}{0.83}       \\ \hline
\end{tabular}
\vspace{-.1in}
\caption{Comparison of UVP for the case $\{\mathbb{P}_{n}\}\subset \mathcal{F}(\mathbb{P}_{0})$,  ${\mathbb{P}_{0}=\mathcal{N}(0, I_{D})}$, $N=4$.
}\vspace{-.1in}
\label{table-gaussians}
\end{table}
% \lingxiao{Why are there two rows for SCRWB?}

\begin{table}[t]
\centering
\scriptsize
\begin{tabular}{|c|c|c|c|c|c|c|c|c|c|}
\hline
\textbf{Metric}                  & \textbf{Method}                & \textbf{D=2}    & \textbf{4}    & \textbf{8}    & \textbf{16}    & \textbf{32}    & \textbf{64}          & \textbf{128}         & \textbf{256}        \\ \hline
\multirow{2}{*}{B$\mathbb{W}_{2}^{2}$-UVP, \%}                  & [FC$\mathbb{W}$B], \cite{cuturi2014fast}               & 0.64   & 0.77    & 1.22   & 3.75    & 8.92    & 14.3          & 18.46         & \color{orange}{21.64}        \\ \cline{2-10}
               & \multirow{2}{*}{[SC$\mathbb{W}_2$B], \citep{fan2020scalable}} & $0.12$ & $0.10$  & $0.19$ & $0.29$  & $0.46$  & $0.6$               & $1.38$              & $2.9$   \\
\cline{1-1}\cline{3-10} 
\multirow{3}{*}{\shortstack{$\mathcal{L}_{2}$-UVP, \%\\(potentials)}} & & $0.17$ & $0.12$  & $0.2$  & $0.31$  & $0.47$  & $0.62$              & $1.21$              & $1.52$              \\
\cline{2-10} 
 &     [CR$\mathbb{W}$B], \citep{li2020continuous}  & $0.58$ & $1.83$  & $8.09$ & \color{orange}{$21.23$} & \color{red}{$55.17$} & \multicolumn{3}{c|}{\color{red}{$>100$}} \\
\cline{2-10} 
 & [C$\mathbb{W}_{2}$B], \underline{\textbf{ours}} & $0.17$ & \color{green}{$0.08$} & \color{green}{$0.06$} & \color{green}{$0.1$}   & \color{green}{$0.2$}   & \color{green}{$0.25$}              & \color{green}{$0.42$}              & \color{green}{$0.82$}              \\
\hline        
\end{tabular}\vspace{-.1in}
\caption{Comparison of UVP for the case $\{\mathbb{P}_{n}\}\subset \mathcal{F}(\mathbb{P}_{0})$,  ${\mathbb{P}_{0}=\text{Uniform}\big([-\sqrt{3},+\sqrt{3}]^{D}}\big)$, $N=4$.
}\vspace{-.1in}
\label{table-rectangles}
\end{table}

In these experiments, our method outperforms [CR$\mathbb{W}$B] and [SC$\mathbb{W}_2$B]. 
% \lingxiao{I think there is no need to bold ``outperform''}
For [CR$\mathbb{W}$B], dimension $\sim\!16$ is the breakpoint: the method does not scale well to higher dimensions.
% \lingxiao{``breakpoint'' is a strange wording. should explain a bit why it won't scale better, possibly due to the uniform regularization measure, where as here the measure is dynamically constructed} 
[SC$\mathbb{W}_2$B] scales with the increasing dimension better, but its errors $\mathcal{L}^{2}$-UVP and B$\mathbb{W}_{2}^{2}$-UVP are twice as high as ours. 
This is likely due to the generative approximation and the difficult min-max-min optimization in [SC$\mathbb{W}_2$B]. 
For completeness, we also compare our algorithm to the proposed in \cite{cuturi2014fast} which approximates the barycenter by a discrete distribution on a fixed number of free-support points. In our experiment, similar to \cite{li2020continuous}, we set $5000$ as the support size. As expected, the B$\mathbb{W}_{2}^{2}$-UVP error of the method increases drastically as the dimension grows and the method is outperformed by our approach.
% \lingxiao{I think more importantly because the optimization is more difficult with min-max-min}.

To show the scalability of our method with the number of input distributions $N$, we conduct an analogous experiment with a high-dimensional location-scatter family for $N=20$. 
We set ${\alpha_{n}=\frac{2n}{N(N+1)}}$ for $n=1,2,...,20$ and choose the uniform distribution on $[-\sqrt{3},+\sqrt{3}]^{D}$ as $\mathbb{P}_{0}$ and construct distributions $\mathbb{P}_{n}\in\mathcal{F}(\mathbb{P}_{0})$ as before. The results for dimensions 32, 64 and 128 are provided in Table \ref{table-rectangles-20}. Similar to the results from Tables \ref{table-gaussians} and \ref{table-rectangles}, we see that our method outperforms the alternatives.

\begin{table}[!h]
\centering
\scriptsize
\begin{tabular}{|c|c|c|c|c|}
\hline
\textbf{Metric}                  & \textbf{Method}                & \textbf{D=32}    & \textbf{64}    & \textbf{128}  \\ \hline
\multirow{2}{*}{B$\mathbb{W}_{2}^{2}$-UVP, \%}                  & [FC$\mathbb{W}$B], \cite{cuturi2014fast}                & 14.09   & \color{orange}{26.21}    & \color{red}{38.43}  \\ \cline{2-5}
               & \multirow{2}{*}{[SC$\mathbb{W}_2$B], \citep{fan2020scalable}} & $0.62$ & $0.93$  & $1.83$  \\
\cline{1-1}\cline{3-5} 
\multirow{2}{*}{\shortstack{$\mathcal{L}_{2}$-UVP, \%\\(potentials)}} & & $0.60$ & $0.86$  & $1.52$      \\
\cline{2-5} 
 & [C$\mathbb{W}_{2}$B], \underline{\textbf{ours}} & \color{green}{$0.31$} & \color{green}{$0.58$} & \color{green}{$1.45$}             \\
\hline        
\end{tabular}\vspace{-.1in}
\caption{Comparison of UVP for the case $\{\mathbb{P}_{n}\}\subset \mathcal{F}(\mathbb{P}_{0})$,  ${\mathbb{P}_{0}=\text{Uniform}\big([-\sqrt{3},+\sqrt{3}]^{D}}\big)$, $N=20$.}
\label{table-rectangles-20}
\end{table}

\subsection{Subset Posterior Aggregation}
\label{sec-subset-posterior}

We apply our method to aggregate subset posterior distributions. % in the Wasserstein-2 space. 
The barycenter of subset posteriors converges to the true posterior \citep{srivastava2018scalable}. Thus, computing the barycenter of subset posteriors is an efficient alternative to obtaining a full posterior in the big data setting \citep{srivastava2015wasp,staib2017parallel,li2020continuous}.

Analogous to \citep{li2020continuous}, we consider Poisson and negative binomial regressions for predicting the hourly number of bike rentals using features such as the day of the week and weather conditions.\footnote{\url{http://archive.ics.uci.edu/ml/datasets/Bike+Sharing+Dataset}} 
We consider the posterior on the 8-dimensional regression coefficients for both Poisson and negative binomial regressions. 
We randomly split the data into $N=5$ equally-sized subsets and obtain $10^{5}$ samples from each subset posterior using the Stan library \citep{carpenter2017stan}.
This gives the discrete uniform distributions $\{\mathbb{P}_{n}\}$ supported on the samples. 
% \lingxiao{mention that each $\mathbb{P}_n$ is an emprical distribution supported on the samples}. 
As the ground truth barycenter $\overline{\mathbb{P}}$, we consider the full dataset posterior also consisting of $10^{5}$ points.

We use B$\mathbb{W}_{2}^{2}$-UVP$(\Tilde{\mathbb{P}},\overline{\mathbb{P}})$ to compare the estimated barycenter $\Tilde{\mathbb{P}}$ (pushforward measure $\nabla\psi^{\dagger}_{n}\sharp\mathbb{P}_{n}$ or generated measure $g\sharp\mathbb{S}$) with the true barycenter. The results are in Table \ref{table-subset-posterior}. All considered methods perform well (UVP$<2\%$), but our method outperforms the alternatives.

\begin{table}[!h]
\centering
\scriptsize
\begin{tabular}{|c|c|c|c|c|c|}
\hline
      & \multirow{2}{*}{Regression} & \multicolumn{2}{c|}{SC$\mathbb{W}_2$B, \citep{fan2020scalable}}    & [CR$\mathbb{W}$B], \citep{li2020continuous}     & C$\mathbb{W}_{2}$B, \underline{\textbf{ours}}     \\ \cline{3-6}
      & & $\Tilde{\mathbb{P}}=g\sharp\mathbb{S}$      & \multicolumn{3}{c|}{$\Tilde{\mathbb{P}}=\nabla\psi_{n}\sharp\mathbb{P}_{n}$} \\ \hline
\multirow{2}{*}{B$\mathbb{W}_{2}^{2}$-UVP, \%} & Poisson &       $0.67$         & $0.41$        & $1.53$    & \color{green}{$0.1$}     \\ \cline{2-6} & negative binomial &       $0.15$         & $0.15$        & $1.26$    & \color{green}{$0.11$}     \\ \hline
\end{tabular}\vspace{-.1in}
\caption{Comparison of UVP for recovered barycenters in our subset posterior aggregation task.}\vspace{-.1in}
\label{table-subset-posterior}
\end{table}

\subsection{Color Palette Averaging}
For qualitative study, we apply our method to aggregating color palettes of images. 
For an RGB image $\mathcal{I}$, its color palette is defined by the discrete uniform distribution $\mathbb{P}(\mathcal{I})$ of all its pixels $\in[0,1]^{3}$. 
For $3$ images $\{\mathcal{I}_{n}\}$ we compute the barycenter $\overline{\mathbb{P}}$ of each color palette $\mathbb{P}_{n}=\mathbb{P}(\mathcal{I}_{n})$ w.r.t.\ uniform weights $\alpha_{n}= \frac{1}{3}$. We apply each computed potential $\nabla\psi^{\dagger}_{n}$ pixel-wise to $\mathcal{I}_{n}$ to obtain the ``pushforward'' image $\nabla\psi^{\dagger}_{n}\sharp \mathcal{I}_{n}$. 
These ``pushforward'' images should be close to the barycenter $\overline{\mathbb{P}}$ of $\{\mathbb{P}_{n}\}$.

\begin{figure}[!h]
     \centering
     \begin{subfigure}[b]{0.6\columnwidth}
         \centering
         \includegraphics[width=\textwidth]{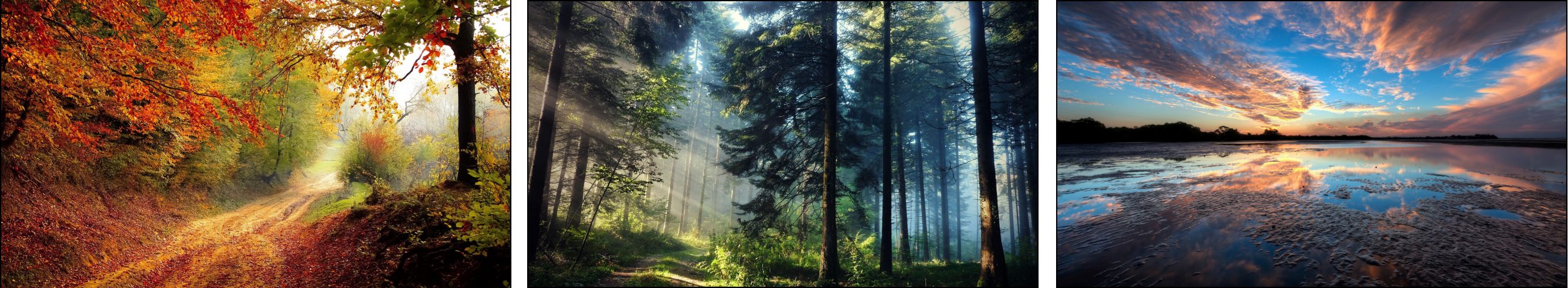}
         \caption{Original images $\{\mathcal{I}_{n}\}$.}
     \end{subfigure}
     \hfill
     \begin{subfigure}[b]{0.38\columnwidth}
         \centering
         \includegraphics[width=\linewidth]{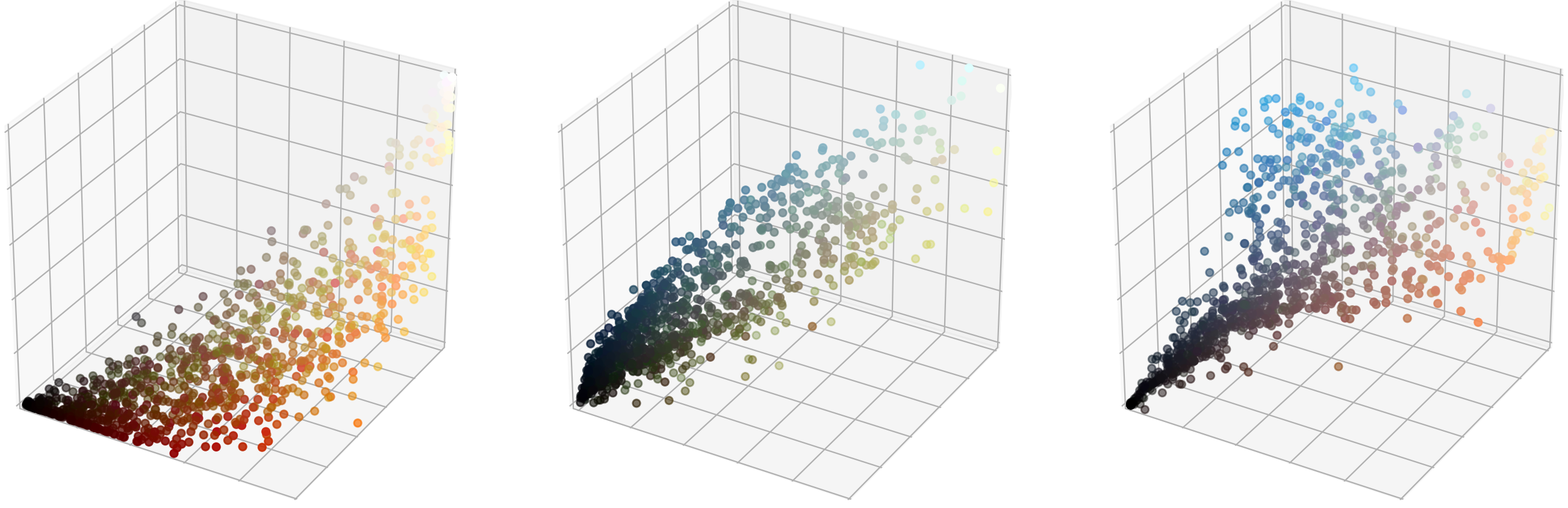}
     \caption{Color palettes $\{\mathbb{P}_{n}\}$ of original images.}
     \end{subfigure}
     
     \vspace{2mm}
     \begin{subfigure}[b]{0.6\columnwidth}
         \centering
         \includegraphics[width=\linewidth]{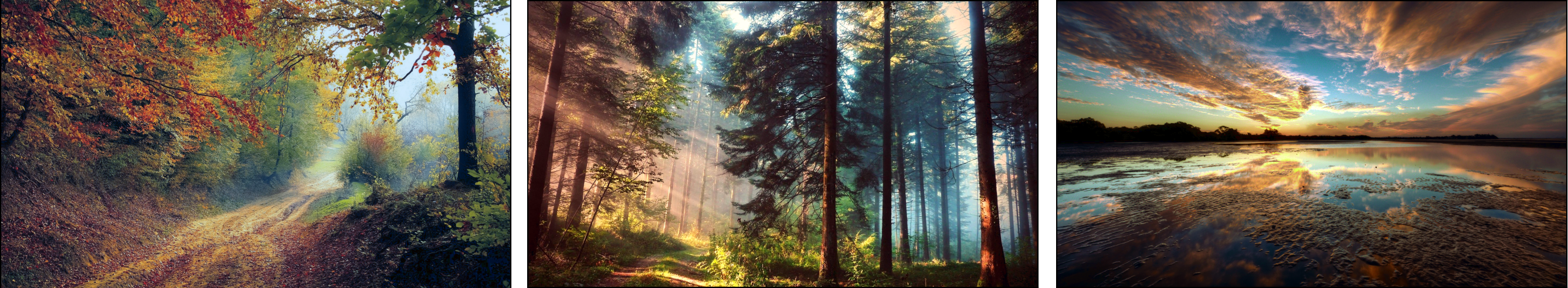}
         \caption{Images with averaged color palette $\{\nabla\psi^{\dagger}_{n}\sharp\mathcal{I}_{n}\}$.}
     \end{subfigure}
     \hfill
     \begin{subfigure}[b]{0.38\columnwidth}
         \centering
         \includegraphics[width=\linewidth]{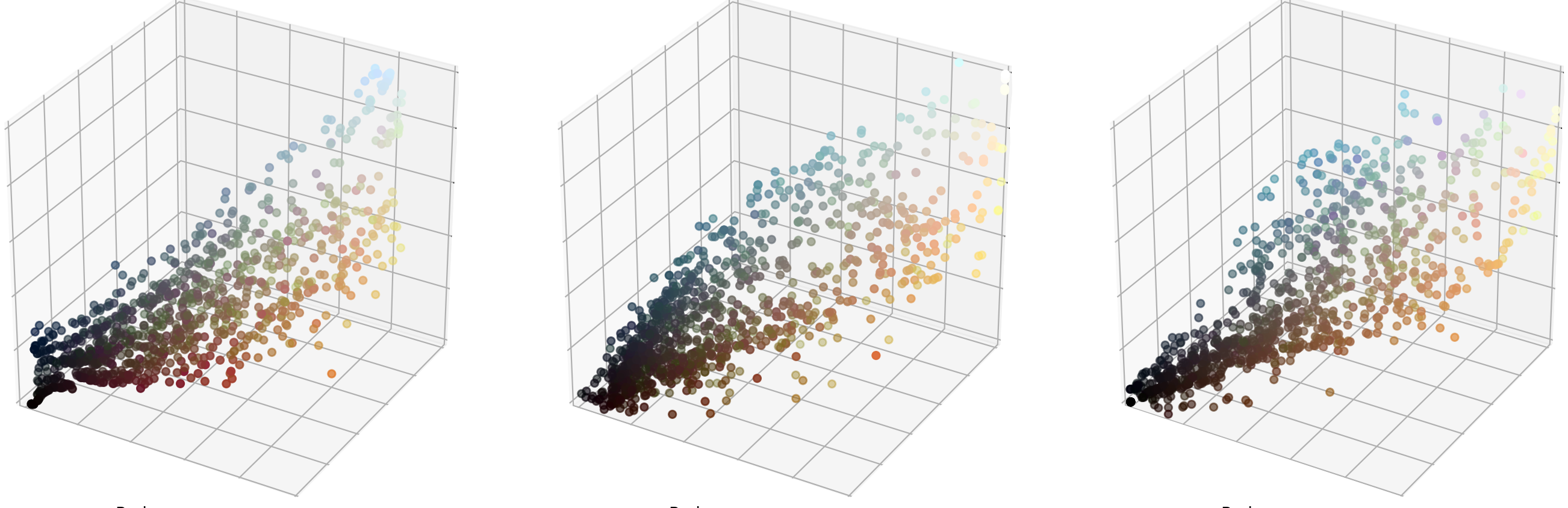}
         \caption{Barycenter palettes $\{\nabla\psi^{\dagger}_{n}\sharp\mathbb{P}_{n}\}$.}
     \end{subfigure}
     \vspace{-.1in}
    \caption{Results of our method applied to averaging color palettes of images.}
    \vspace{-.1in}
    \label{fig:images-palettes}
\end{figure}
The results are provided in Figure \ref{fig:images-palettes}. 
Note that the image $\nabla\psi^{\dagger}_{1}\sharp \mathcal{I}_{1}$ inherits certain attributes of images $\mathcal{I}_{2}$ and $\mathcal{I}_{3}$: the sky becomes bluer and the trees becomes greener. 
% This observation is consistent with the fact that $\mathcal{I}_{2},\mathcal{I}_{3}$ have a lot of blue and green color respectively. 
On the other hand, the sunlight in images $\nabla\psi^{\dagger}_{2}\sharp \mathcal{I}_{2},\nabla\psi^{\dagger}_{3}\sharp \mathcal{I}_{3}$ has acquired an orange tint, thanks to the dominance of orange in $\mathcal{I}_{1}$.

\subsubsection*{Acknowledgments}
The Skoltech Advanced Data Analytics in Science and Engineering Group thanks the Skoltech CDISE HPC Zhores cluster staff for computing cluster provision and Skoltech-MIT NGP initiative for the support.

The MIT Geometric Data Processing group acknowledges the generous support of Army Research Office grant W911NF2010168, of Air Force Office of Scientific Research award FA9550-19-1-031, of National Science Foundation grant IIS-1838071, from the CSAIL Systems that Learn program, from the MIT–IBM Watson AI Laboratory, from the Toyota--CSAIL Joint Research Center, from a gift from Adobe Systems, from an MIT.nano Immersion Lab/NCSOFT Gaming Program seed grant, and from the Skoltech--MIT Next Generation Program.
% Use unnumbered third level headings for the acknowledgments. All
% acknowledgments, including those to funding agencies, go at the end of the paper.

\bibliography{references}
\bibliographystyle{iclr2021_conference}

\clearpage
\appendix
\section{The Algorithm}
\label{sec-algorithm-procedure}

The numerical procedure for solving our final objective \eqref{main-objective} is given below.

\begin{algorithm}
\SetAlgorithmName{Algorithm}{empty}{Empty}
\SetKwInOut{Input}{Input}
\Input{Distributions $\mathbb{P}_{1},\dots,\mathbb{P}_{N}$ with sample access;\\Weights $\alpha_{1},\dots,\alpha_{N}\geq 0$ with $\sum_{n=1}^{N}\alpha_{n}=1$\;
Regularization distribution $\widehat{\mathbb{P}}'$ given by a sampler\; 
Congruence regularizer coefficient $\tau\geq 1$\;
Balancing coefficient $\gamma\in [0,1]$\;
Cycle-consistency regularizer coefficient $\lambda>0$\;
$2N$ ICNNs $\{\psi_{\theta_{n}},\overline{\psi_{\omega_{n}}}\}$\;Batch size $K>0$\;}

\For{$t=1,2,\dots$}{
  1. Sample batches $X_{n}\sim \mathbb{P}_{n}$ for all $n=1,\dots,N$\;
  2. Compute the pushforwards $Y_{n}=\nabla\psi_{\theta_{n}}\sharp X_{n}$  for all $n=1,\dots,N$\;
  3. Sample batch $Y_{0}\sim \widehat{\mathbb{P}}$\;
  4. Compute the Monte-Carlo estimate of the congruence regularizer:
  $$\mathcal{L}_{\text{Congruence}}:=\frac{1}{K}\cdot\sum_{n=1}^{N}\gamma_{n}\sum_{y\in Y_{n}}\big[\sum_{n'=1}^{N}\alpha_{n'}\overline{\psi_{\omega_{n'}}}(y)-\frac{\|y\|^{2}}{2}\big]_{+},$$
  where $\gamma_{0}=\gamma$ and $\gamma_{n}=\alpha_{n}\cdot (1-\gamma)$ for $n=1,2,\dots,N$\;
  5. Compute  the Monte-Carlo estimate of the cycle-consistency regularizer:
  $$\mathcal{L}_{\text{Cycle}}:=\frac{1}{K}\sum_{n=1}^{N}\alpha_{n}\bigg[\sum_{x\in X_{n}}\|\nabla\overline{\psi_{\omega_{n}}}\circ\nabla\psi_{\theta_{n}}(x)-x\|_{2}^{2}\bigg];$$
  6. Compute  the Monte-Carlo estimate of multiple correlations:
  $$\mathcal{L}_{\text{MultiCorr}}:=\sum_{n=1}^{N}\bigg[\alpha_{n}\cdot \frac{1}{K}\sum_{x\in X_{n}}\big[\langle x,\nabla\psi_{\theta_{n}}(x)\rangle-\overline{\psi_{\omega_{n}}}(\nabla\psi_{\theta_{n}}(x))]\bigg];$$
  7. Compute the total loss:
  $$\mathcal{L}_{\text{Total}}:=\mathcal{L}_{\text{MultiCorr}}+\lambda\cdot\mathcal{L}_{\text{Cycle}}+\tau\cdot\mathcal{L}_{\text{Congruence}};$$
  8. Perform a gradient step over $\{\theta_{n},\omega_{n}\}$ by using $\frac{\partial \mathcal{L}_{\text{Total}}}{\partial \{\theta_{n},\omega_{n}\}}$\;
 }
\caption{Numerical Procedure for Optimizing Multiple Correlations \eqref{main-objective}}
\label{algorithm-main}
\end{algorithm}

%The optimization is done via mini-batch stochastic gradient descent by using Monte-Carlo estimates by random samples from $\{\mathbb{P}_{n}\}$ and $\hat{\mathbb{P}}$.

\textbf{Parametrization of the potentials.} To parametrize potentials $\{\psi_{\theta_{n}},\overline{\psi_{\omega_{n}}}\}$, we use DenseICNN (dense input convex neural network) with quadratic skip connections; see \cite[Appendix B.2]{korotin2019wasserstein}. 
As an initialization step, we pre-train the potentials to satisfy 
$$\psi_{\theta_{n}}(x)\approx\frac{\|x\|^{2}}{2}\qquad\text{and}\qquad\overline{\psi_{\omega_{n}}}(y)\approx\frac{\|y\|^{2}}{2}.$$
Such pre-training provides a good start for the networks: each $\psi_{\theta_{n}}$ is approximately conjugate to the corresponding $\overline{\psi_{\omega_{n}}}$.
On the other hand, the initial networks $\{\psi_{\theta_{n}}\}$ are approximate congruent according to \eqref{bar-char}.

\textbf{Computational Complexity.} For a single training iteration, the time complexity of both forward (evaluation) and backward (computing the gradient with respect to the parameters) passes through the objective function  \eqref{main-objective} is $O(NT)$. Here $N$ is the number of input distributions and $T$ is the time taken by evaluating each individual potential (parameterized as a neural network) on a batch of points sampled from either $\mathbb{P}_n$ or $\widehat{\mathbb{P}}$. This claim follows from the well-known fact that gradient evaluation $\nabla_\theta h_\theta(x)$ of $h_\theta: \mathbb{R}^{D} \to \mathbb{R}$, when parameterized as a neural network, requires time proportional to the size of the computational graph. Hence, gradient computation requires computational time proportional to the time for evaluating the function $h_\theta(x)$ itself. The same holds when computing the derivative with respect to $x$. Then, for instance, computing the term $\nabla\overline{\psi_{n}^{\ddagger}}\circ \nabla\psi_{n}^{\dagger} (x)$ in (14) takes $O(T)$ time. The gradient of this term with respect to $\theta$ also takes $O(T)$ time: Hessian-vector products that appear can be calculated in $O(T)$ time using the famous Hessian trick, see \cite{pearlmutter1994fast}. 

In practice, we compute all the gradients using automatic differentiation. We empirically measured that for our DenseICNN potentials, the computation of their gradient w.r.t. input $x$, i.e., $\nabla \psi^\dagger(x)$, requires roughly 3-4x more time than the computation of $\psi^\dagger(x)$.

\section{Proofs}
\label{sec-proofs}
In this section, we prove our main Theorems \ref{thm-main} and \ref{thm-main-top}.

We use $\mathcal{L}^{2}(\mathbb{R}^{D}\rightarrow \mathbb{R}^{D},\mu)$ to denote the \textbf{Hilbert space} of functions ${f:\mathbb{R}^{D}\rightarrow\mathbb{R}^{D}}$ with integrable square w.r.t. a probability measure $\mu$. The corresponding inner product for $f_{1},f_{2}\in \mathcal{L}^{2}(\mathbb{R}^{D}\rightarrow \mathbb{R}^{D}, \mu)$ is denoted by 
$$\langle f_{1},f_{2}\rangle_{\mu}\stackrel{\text{def}}{=}\int_{\mathbb{R}^{D}}\langle f_{1}(x),f_{2}(x)\rangle d\mu(x),$$
where $\langle f_1(x), f_2(x) \rangle$ is the Euclidean dot product.
% \lingxiao{should probably mention $\langle f_1(x), f_2(x) \rangle$ is just Euclidean dot product}
We use $\|\cdot\|_{\mu}=\sqrt{\langle\cdot,\cdot\rangle_{\mu}}$ to denote the norm induced by the inner product in $\mathcal{L}^{2}(\mathbb{R}^{D}\rightarrow \mathbb{R}^{D},\mu)$.

We also recall a useful property of lower semi-continuous convex function $\psi:\mathbb{R}^{D}\rightarrow \mathbb{R}$:
\begin{equation}\nabla \psi(x) = \argmax_{y\in\mathbb{R}^{D}} \big[\langle y, x \rangle - \overline{\psi}(y)\big]
\label{lsc-grad},
\end{equation}
which follows from the fact that $$\hat{y} = \argmax_{y\in\mathbb{R}^{D}} \big[\langle y, x \rangle - \overline{\psi}(y)\big]\iff x-\nabla\overline{\psi}(\hat{y})=0.$$

% \lingxiao{The theorem should be restated here for easy reference}
We begin with the proof of Theorem \ref{thm-main}.

\begin{proof}We consider the difference between the estimated correlations and true ones:
\begin{eqnarray}
\Delta=\sum_{n=1}^{N}\alpha_{n}\int_{\mathbb{R}^{D}}\psi_{n}(x)d\mathbb{P}_{n}(x)-\sum_{n=1}^{N}\alpha_{n}\int_{\mathbb{R}^{D}}\psi_{n}^{*}(x)d\mathbb{P}_{n}(x)=
\nonumber
\\
\sum_{n=1}^{N}\alpha_{n}\int_{\mathbb{R}^{D}}\big[\langle\nabla\psi_{n}(x),x\rangle-\overline{\psi_{n}}\big(\nabla\psi_{n}(x))\big]d\mathbb{P}_{n}(x)-
\nonumber
\\
\sum_{n=1}^{N}\alpha_{n}\int_{\mathbb{R}^{D}}\big[\langle\nabla\psi_{n}^{*}(x),x\rangle-\overline{\psi_{n}^{*}}\big(\nabla\psi_{n}^{*}(x))\big]d\mathbb{P}_{n}(x),
\label{main-repr}
\end{eqnarray}
where we twice use \eqref{lsc-grad} for $f=\psi_{n}$ and $f=\psi^{*}_{n}$.
We note that
\begin{eqnarray}
\sum_{n=1}^{N}\alpha_{n}\int_{\mathbb{R}^{D}}\langle\nabla\psi_{n}^{*}(x),x\rangle d\mathbb{P}_{n}(x)=\sum_{n=1}^{N}\alpha_{n}\int_{\mathbb{R}^{D}}\langle y,\nabla\overline{\psi_{n}^{*}}(y)\rangle d\overline{\mathbb{P}}(y)=
\nonumber
\\
\int_{\mathbb{R}^{D}}\langle y,\sum_{n=1}^{N}\alpha_{n}\nabla\overline{\psi_{n}^{*}}(y)\rangle d\overline{\mathbb{P}}(y)=\int_{\mathbb{R}^{D}}\langle y,y\rangle d\overline{\mathbb{P}}(y)=\|\text{id}_{\mathbb{R}^{D}}\|_{\overline{\mathbb{P}}}^{2},
\label{third-summand}
\end{eqnarray}
where we use of change-of-variable formula for $\nabla \psi_n^{*}\sharp\mathbb{P}_n = \overline{\mathbb{P}}$ and \eqref{bar-char}.
% \lingxiao{mention the use of change-of-variable for $(\nabla \psi_n^*)_\sharp \mathbb{P}_n = \overline{\mathbb{P}}$ and the use of Eq (5)}
Analogously,
\begin{eqnarray}
\sum_{n=1}^{N}\alpha_{n}\int_{\mathbb{R}^{D}}\overline{\psi_{n}^{*}}\big(\nabla\psi_{n}^{*}(x))d\mathbb{P}_{n}(x)=\sum_{n=1}^{N}\alpha_{n}\int_{\mathbb{R}^{D}}\overline{\psi_{n}^{*}}\big(y)d\overline{\mathbb{P}}(y)=
\nonumber
\\
\int_{\mathbb{R}^{D}}\sum_{n=1}^{N}\alpha_{n}\overline{\psi_{n}^{*}}\big(y)d\overline{\mathbb{P}}(y)=\int_{\mathbb{R}^{D}}\frac{\|y\|^{2}}{2}d\overline{\mathbb{P}}(y)=\frac{1}{2}\|\text{id}_{\mathbb{R}^{D}}\|_{\overline{\mathbb{P}}}^{2}.
\label{forth-summand}
\end{eqnarray}
% \lingxiao{Comment that $MC(\{\psi_n^*\}) = \frac{1}{2}\norm{\text{id}}_{\overline{\mathbb{P}}}^2$}
Since each $\psi_{n}$ is $\mathcal{B}$-smooth, we conclude that $\overline{\psi_{n}}$ is $\frac{1}{\mathcal{B}}$-strongly convex, see \citep{kakade2009duality}.
% \lingxiao{cite this result, e.g., Kakade 09}
Thus, we have
\begin{eqnarray}\overline{\psi_{n}}\big(\nabla\psi_{n}^{*}(x)))\geq
\nonumber
\\
\overline{\psi_{n}}\big(\nabla\psi_{n}(x)))+\langle \underbrace{\nabla \overline{\psi_n^\dagger}\circ\nabla \psi_n^\dagger (x)}_{=x},\nabla\psi_{n}^{*}(x)-\nabla\psi_{n}(x)\rangle+\frac{1}{2\mathcal{B}}\|\nabla\psi_{n}^{*}(x)-\nabla\psi_{n}(x)\|^{2}=
\nonumber
\\
\overline{\psi_{n}}\big(\nabla\psi_{n}(x)))+\langle x,\nabla\psi_{n}^{*}(x)-\nabla\psi_{n}(x)\rangle+\frac{1}{2\mathcal{B}}\|\nabla\psi_{n}^{*}(x)-\nabla\psi_{n}(x)\|^{2},
\end{eqnarray}
% \lingxiao{mention the identity $\nabla \overline{\psi_n^\dagger}(\nabla \psi_n^\dagger (x)) = x$ used in the second term}
or equivalently
\begin{equation}-\overline{\psi_{n}}\big(\nabla\psi_{n}(x)))\geq -\overline{\psi_{n}}\big(\nabla\psi_{n}^{*}(x)))+\langle x,\nabla\psi_{n}^{*}(x)-\nabla\psi_{n}(x)\rangle+\frac{1}{2\mathcal{B}}\|\nabla\psi_{n}^{*}(x)-\nabla\psi_{n}(x)\|^{2}.
\label{str-conv-non-int}
\end{equation}
We integrate \eqref{str-conv-non-int} w.r.t. $\mathbb{P}_{n}$ and sum over $n=1,2,\dots,N$ with weights $\alpha_{n}$:
\begin{eqnarray}
-\sum_{n=1}^{N}\alpha_{n}\int_{\mathbb{R}^{D}}\overline{\psi_{n}}\big(\nabla\psi_{n}(x))d\mathbb{P}_{n}(x)\geq
\nonumber
\\
-\sum_{n=1}^{N}\alpha_{n}\int_{\mathbb{R}^{D}}\overline{\psi_{n}}\big(\nabla\psi_{n}^{*}(x))d\mathbb{P}_{n}(x)+\sum_{n=1}^{N}\alpha_{n}\langle x,\nabla\psi_{n}^{*}(x)\rangle_{\mathbb{P}_{n}}-\sum_{n=1}^{N}\alpha_{n}\langle x,\nabla\psi_{n}(x)\rangle_{\mathbb{P}_{n}}+
\nonumber
\\
\sum_{n=1}^{N}\alpha_{n}\frac{1}{2\mathcal{B}}\|\nabla\psi_{n}^{*}(x)-\nabla\psi_{n}(x)\|^{2}_{\mathbb{P}_{n}}=
\nonumber
\\
-\int_{\mathbb{R}^{D}}\sum_{n=1}^{N}\alpha_{n}\overline{\psi_{n}}\big(y)d\overline{\mathbb{P}}(y)+\sum_{n=1}^{N}\alpha_{n}\langle x,\nabla\psi_{n}^{*}(x)\rangle_{\mathbb{P}_{n}}-
\nonumber
\\
\sum_{n=1}^{N}\alpha_{n}\langle x,\nabla\psi_{n}(x)\rangle_{\mathbb{P}_{n}}+
\sum_{n=1}^{N}\alpha_{n}\frac{1}{2\mathcal{B}}\|\nabla\psi_{n}^{*}(x)-\nabla\psi_{n}(x)\|^{2}_{\mathbb{P}_{n}}.
\label{first-part1}
% \\
% \|\text{id}_{\mathbb{R}^{D}}\|_{\overline{\mathbb{P}}}^{2}-\frac{1}{2}\|\text{id}_{\mathbb{R}^{D}}\|_{\overline{\mathbb{P}}}^{2}-\sum_{n=1}^{N}\alpha_{n}\langle x,\nabla\psi_{n}(x)\rangle_{\mathbb{P}_{n}}+\sum_{n=1}^{N}\alpha_{n}\frac{1}{2\mathcal{B}}\|\nabla\psi_{n}^{*}(x)-\nabla\psi_{n}(x)\|^{2}_{\mathbb{P}_{n}}
% \nonumber
\end{eqnarray}
We note that
\begin{eqnarray}
-\int_{\mathbb{R}^{D}}\sum_{n=1}^{N}\alpha_{n}\overline{\psi_{n}}\big(y)d\overline{\mathbb{P}}(y)=\int_{\mathbb{R}^{D}}\big[\frac{\|y\|^{2}}{2}-\sum_{n=1}^{N}\alpha_{n}\overline{\psi_{n}}\big(y)\big]d\overline{\mathbb{P}}(y)-\int_{\mathbb{R}^{D}}\frac{\|y\|^{2}}{2}d\overline{\mathbb{P}}(y)
\nonumber
\\
\int_{\mathbb{R}^{D}}\big[\frac{\|y\|^{2}}{2}-\sum_{n=1}^{N}\alpha_{n}\overline{\psi_{n}}\big(y)\big]d\overline{\mathbb{P}}(y)-\frac{1}{2}\|\text{id}_{\mathbb{R}^{D}}\|_{\overline{\mathbb{P}}}^{2}.
\label{first-part-2}
\end{eqnarray}
% \lingxiao{I worked it out. You should probably mention that the term $\sum_{n=1}^N \alpha_n \langle x, \nabla\psi_n^*(x)\rangle_{\mathbb{P}_n} = \norm{\text{id}}^2_{\overline{\mathbb{P}}}$, so it is canceled by the last term in (24) and $MC(\{\psi_n^*\})$}
Now we substitute \eqref{first-part1}, \eqref{first-part-2}, \eqref{third-summand} and \eqref{forth-summand} into \eqref{main-repr} to obtain \eqref{cong-generative}.
% $$\Delta\geq \frac{1}{2\mathcal{B}}\sum_{n=1}^{N}\alpha_{n}\|\nabla\psi_{n}^{*}(x)-\nabla\psi_{n}(x)\|^{2}_{\mathbb{P}_{n}}\geq \frac{1}{2\mathcal{B}}\sum_{n=1}^{N}\alpha_{n}\mathbb{W}_{2}^{2}\big(\nabla\psi_{n}\circ\mathbb{P}_{n}, \overline{\mathbb{P}}).$$
\end{proof}

Next, we prove Theorem \ref{thm-main-top}.

\begin{proof}
Since $\overline{\psi_{n}^{\ddagger}}$ is $\beta^{\ddagger}$ strongly convex, its conjugate $\psi_{n}^{\ddagger}$ is $\frac{1}{\beta^{\ddagger}}$-smooth, i.e. has $\frac{1}{\beta^\ddagger}$-Lipschitz gradient 
% \lingxiao{should be $\frac{1}{\beta^\ddagger}$-Lipschitz}
$\nabla \psi_{n}^{\ddagger}$ \citep{kakade2009duality}. 
Thus, for all $x,x'\in\mathbb{R}^{D}$:
$$\|\nabla \psi_{n}^{\ddagger}(x)-\nabla \psi_{n}^{\ddagger}(x')\|^{2}\leq (\frac{1}{\beta^{\ddagger}})^{2}\cdot \|x-x'\|^{2}.$$
We substitute $x'=\nabla \overline{\psi_{n}^{\ddagger}}\circ\nabla\psi_{n}^{\dagger}(y)=\big(\nabla \psi_{n}^{\ddagger}\big)^{-1}\circ\nabla\psi_{n}^{\dagger}(y)$ and obtain:
\begin{equation}\|\nabla\psi_{n}^{\dagger}(x)-\nabla\psi_{n}^{\ddagger}(x)\|^{2}\leq(\frac{1}{\beta^{\ddagger}})^{2}\|x-\nabla\overline{\psi_{n}^{\ddagger}}\circ \nabla\psi_{n}^{\dagger}(x)\|^{2}.
\label{reg-map-bound}
\end{equation}
Since the function $\overline{\psi_n^\ddagger}$ is $\mathcal{B}^{\ddagger}$-smooth, we have for all $x\in\mathbb{R}^{D}$:
% \lingxiao{notation: $\mathcal{B}^\ddagger$ instead of $\mathcal{B}^\dagger$}
\begin{eqnarray}
\overline{\psi_{n}^{\ddagger}}(\nabla\psi_{n}^{\dagger}(x))\leq \overline{\psi_{n}^{\ddagger}}(\nabla\psi_{n}^{\ddagger}(x))+\langle \underbrace{\nabla\overline{\psi_{n}^{\ddagger}}\circ \nabla\psi_{n}^{\ddagger}(x)}_{=x}, \nabla\psi_{n}^{\dagger}(x)-\nabla\psi_{n}^{\ddagger}(x)\rangle+\frac{\mathcal{B}^{\ddagger}}{2}\|\nabla\psi_{n}^{\dagger}(x)-\nabla\psi_{n}^{\ddagger}(x)\|^{2},
\nonumber
\end{eqnarray}
that is equivalent to:
\begin{eqnarray}
\langle x, \nabla\psi_{n}^{\dagger}(x)\rangle-\overline{\psi_{n}^{\ddagger}}(\nabla\psi_{n}^{\dagger}(x))\geq \underbrace{\langle x, \nabla\psi_{n}^{\ddagger}(x)\rangle-\overline{\psi_{n}^{\ddagger}}(\nabla\psi_{n}^{\ddagger}(x))}_{\psi^{\ddagger}_{n}(x)}-\frac{\mathcal{B}^{\ddagger}}{2}\|\nabla\psi_{n}^{\dagger}(x)-\nabla\psi_{n}^{\ddagger}(x)\|^{2}.
\label{smoothness-ineq}
\end{eqnarray}
We combine \eqref{smoothness-ineq} with \eqref{reg-map-bound} to obtain
\begin{equation}
\langle x, \nabla\psi_{n}^{\dagger}(x)\rangle-\overline{\psi_{n}^{\ddagger}}(\nabla\psi_{n}^{\dagger}(x))\geq \psi^{\ddagger}_{n}(x)-\frac{\mathcal{B}^{\ddagger}}{2(\beta^{\ddagger})^{2}}\cdot\|\text{id}_{\mathbb{R}^{D}}-\nabla\overline{\psi_{n}^{\ddagger}}\circ \nabla\psi_{n}^{\dagger}\|^{2}.
\label{smoothness-ineq-short}
\end{equation}

For every $n=1,2,\dots,N$ we integrate \eqref{smoothness-ineq-short} w.r.t. $\mathbb{P}_{n}$ and sum up the corresponding cycle-consistency regularization term:
% \lingxiao{mention that you additionally add the $\lambda$ term on both sides}
\begin{eqnarray}\int_{\mathbb{R}^{D}}\big[\langle x,\nabla\psi_{n}^{\dagger}(x)\rangle-\overline{\psi_{n}^{\ddagger}}(\nabla\psi_{n}^{\dagger}(x))]d\mathbb{P}_{n}(x)+\lambda\cdot \|\nabla\overline{\psi_{n}^{\ddagger}}\circ \nabla\psi_{n}^{\dagger}-\text{id}_{\mathbb{R}^{D}}\|_{\mathbb{P}_{n}}^{2}\geq 
\nonumber
\\
\int_{\mathbb{R}^{D}}\psi^{\ddagger}(x)d\mathbb{P}_{n}(x)+\big(\lambda-\frac{\mathcal{B}^{\dagger}}{2(\beta^{\ddagger})^{2}}\big)\cdot \underbrace{\|\nabla\overline{\psi_{n}^{\ddagger}}\circ \nabla\psi_{n}^{\dagger}-\text{id}_{\mathbb{R}^{D}}\|_{\mathbb{P}_{n}}^{2}}_{\mathcal{R}_{2}^{\mathbb{P}_{n}}(\psi_{n}^{\dagger}, \overline{\psi_{n}^{\ddagger}})}.
\label{creg-corr-part-lower}
\end{eqnarray}
We sum \eqref{creg-corr-part-lower} for $n=1,2,\dots,N$ w.r.t. weights $\alpha_{n}$ to obtain:
\begin{eqnarray}
\sum_{n=1}^{N}\alpha_{n}\int_{\mathbb{R}^{D}}\big[\langle x,\nabla\psi_{n}^{\dagger}(x)\rangle-\overline{\psi_{n}^{\ddagger}}(\nabla\psi_{n}^{\dagger}(x))]d\mathbb{P}_{n}(x)+\lambda\sum_{n=1}^{N}\alpha_{n} \mathcal{R}_{2}^{\mathbb{P}_{n}}(\psi_{n}^{\dagger}, \overline{\psi_{n}^{\ddagger}})\geq \nonumber
\\
\underbrace{\sum_{n=1}^{N}\alpha_{n}\int_{\mathbb{R}^{D}}\psi^{\ddagger}(x)d\mathbb{P}_{n}(x)}_{\text{\normalfont MultiCorr}(\{\alpha_{n},\mathbb{P}_{n}\}\mid\{\psi^{\ddagger}_{n}\})}+\sum_{n=1}^{N}\alpha_{n}\big(\lambda-\frac{\mathcal{B}^{\dagger}}{2(\beta^{\ddagger})^{2}}\big)\cdot \mathcal{R}_{2}^{\mathbb{P}_{n}}(\psi_{n}^{\dagger}, \overline{\psi_{n}^{\ddagger}}).
\nonumber
% \\
% \geq \text{\normalfont MultiCorr}(\{\mathbb{P}_{n}\}\mid\{\psi^{*}_{n}\})-
% \nonumber
% \\
% \bigg[\int_{\mathbb{R}^{D}}\sum_{n=1}^{N}\alpha_{n}\overline{\psi_{n}^{\ddagger}}\big(y)d\overline{\mathbb{P}}(x)-\frac{1}{2}\|\text{id}_{\mathbb{R}^{D}}\|_{\overline{\mathbb{P}}}^{2}\bigg]+\frac{1}{2\mathcal{B}^{\dagger}}\sum_{n=1}^{N}\alpha_{n}\|\nabla\psi_{n}^{*}(x)-\nabla\psi_{n}^{\ddagger}(x)\|^{2}_{\mathbb{P}_{n}}.
\label{half-bound}
\end{eqnarray}
We add $\tau\cdot\mathcal{R}_{1}^{\widehat{{\mathbb{P}}}}(\{\overline{\psi_{n}^{\ddagger}}\})$ to both sides of \eqref{half-bound} to get
\begin{eqnarray}
\text{MultiCorr}\big(\{\alpha_{n},\mathbb{P}_{n}\}\mbox{ }\vert\mbox{ } \{\psi_{n}^{\dagger}\}, \{\overline{\psi_{n}^{\ddagger}}\}; \tau,\widehat{{\mathbb{P}}}, \lambda\big)\geq \text{\normalfont MultiCorr}(\{\alpha_{n},\mathbb{P}_{n}\}\mid\{\psi^{\ddagger}_{n}\})+
\nonumber
\\
\tau\cdot\mathcal{R}_{1}^{\widehat{{\mathbb{P}}}}(\{\overline{\psi_{n}^{\ddagger}}\})+\sum_{n=1}^{N}\alpha_{n}\big(\lambda-\frac{\mathcal{B}^{\dagger}}{2(\beta^{\ddagger})^{2}}\big)\cdot \mathcal{R}_{2}^{\mathbb{P}_{n}}(\psi_{n}^{\dagger}, \overline{\psi_{n}^{\ddagger}}).
% \bigg[\int_{\mathbb{R}^{D}}\sum_{n=1}^{N}\alpha_{n}\overline{\psi_{n}^{\ddagger}}\big(y)d\overline{\mathbb{P}}(x)-\frac{1}{2}\|\text{id}_{\mathbb{R}^{D}}\|_{\overline{\mathbb{P}}}^{2}\bigg]+\frac{1}{2\mathcal{B}^{\dagger}}\sum_{n=1}^{N}\alpha_{n}\|\nabla\psi_{n}^{*}(x)-\nabla\psi_{n}^{\ddagger}(x)\|^{2}_{\mathbb{P}_{n}}\geq
% \nonumber
% \\
% \frac{1}{2}^{\dagger}}\sum_{n=1}^{N}\alpha_{n}\|\nabla\psi_{n}^{*}(x)-\nabla\psi_{n}^{\ddagger}(x)\|^{2}_{\mathbb{P}_{n}}\geq 
\label{semi-final-bound}
\end{eqnarray}
% By combining \eqref{eps-def-thm2} with the left-hand-side of \eqref{semi-final-bound} and Theorem \ref{thm-main} with the right-had-side we derive \lingxiao{better to write we subtract $\text{MultiCorr}(\psi_n^*)$ from both sides}:
We substract $\text{\normalfont MultiCorr}(\{\alpha_{n},\mathbb{P}_{n}\}\mid\{\psi^{\ddagger}_{n}\})$ from both sides and use Theorem \ref{thm-main} to obtain
\begin{eqnarray}
\Delta\geq-\int_{\mathbb{R}^{D}}\sum_{n=1}^{N}\big[\alpha_{n}\overline{\psi_{n}^{\ddagger}}(y)-\frac{\|y\|^{2}}{2}\big]d\overline{\mathbb{P}}(y)+\frac{\beta^{\ddagger}}{2}\sum_{n=1}^{N}\alpha_{n}\|\nabla\psi_{n}^{*}(x)-\nabla\psi_{n}^{\ddagger}(x)\|^{2}_{\mathbb{P}_{n}}+
\label{eps-lower-first}
\\
\tau\cdot\mathcal{R}_{1}^{\widehat{{\mathbb{P}}}}(\{\overline{\psi_{n}^{\ddagger}}\})+
\sum_{n=1}^{N}\alpha_{n}\big(\lambda-\frac{\mathcal{B}^{\dagger}}{2(\beta^{\ddagger})^{2}}\big)\cdot \mathcal{R}_{2}^{\mathbb{P}_{n}}(\psi_{n}^{\dagger}, \overline{\psi_{n}^{\ddagger}})\geq
\label{eps-lower-second}
\\
\sum_{n=1}^{N}\alpha_{n}\big(\lambda-\frac{\mathcal{B}^{\dagger}}{2(\beta^{\ddagger})^{2}}\big)\cdot \mathcal{R}_{2}^{\mathbb{P}_{n}}(\psi_{n}^{\dagger}, \overline{\psi_{n}^{\ddagger}})+\frac{\beta^{\ddagger}}{2}\sum_{n=1}^{N}\alpha_{n}\|\nabla\psi_{n}^{*}(x)-\nabla\psi_{n}^{\ddagger}(x)\|^{2}_{\mathbb{P}_{n}}.
\label{eps-lower}
\end{eqnarray}
% \lingxiao{Mention the first term and the $\mathcal{R}_1^{\hat{\mathbb{P}}}$ term result in a nonnegativity term, based on the assumptions}
In transition from \eqref{eps-lower-second} to \eqref{eps-lower}, we explot the fact that the sum of the first term of \eqref{eps-lower-first} with the regularizer $\tau\cdot\mathcal{R}_{1}^{\widehat{{\mathbb{P}}}}(\{\overline{\psi_{n}^{\ddagger}}\})$. Since $\lambda>\frac{\mathcal{B}^{\dagger}}{2(\beta^{\ddagger})^{2}}$, from \eqref{eps-lower} we immediately conclude $\Delta\geq 0$; i.e., the \textbf{multiple correlations upper bound} \eqref{multicorr-upper-bound} holds true. 
On the other hand, for every $n=1,2,\dots,N$ we have
\begin{eqnarray}\|\nabla\psi_{n}^{*}(x)-\nabla\psi_{n}^{\ddagger}(x)\|^{2}_{\mathbb{P}_{n}}\leq \frac{2\Delta}{\alpha_{n}\beta^{\ddagger}}\quad\text{and}\quad \|\nabla\overline{\psi_{n}^{\ddagger}}\circ \nabla\psi_{n}^{\dagger}-\text{id}_{\mathbb{R}^{D}}\|_{\mathbb{P}_{n}}^{2}\leq \frac{2\Delta}{\alpha_{n}\cdot(\lambda-\frac{\mathcal{B}^{\dagger}}{2(\beta^{\ddagger})^{2}})}.
\label{two-conclusions}
\end{eqnarray}
We combine the second part of \eqref{two-conclusions} with \eqref{reg-map-bound} integrated w.r.t. $\mathbb{P}_{n}$:
% \lingxiao{this sentense is problematic}
\begin{equation}
\|\nabla\psi_{n}^{\ddagger}-\nabla\psi_{n}^{\dagger}\|_{\mathbb{P}_{n}}^{2}\leq \frac{2\Delta}{\alpha_{n}\cdot(\lambda(\beta^{\ddagger})^{2}-\frac{\mathcal{B}^{\dagger}}{2})}.
\label{diff-grad-potential-bound}
\end{equation}
Finally, we use the triangle inequality for $\|\cdot\|_{\mathbb{P}_{n}}$ and conclude
\begin{eqnarray}\|\nabla\psi_{n}^{*}-\nabla\psi_{n}^{\dagger}\|_{\mathbb{P}_{n}}\leq \|\nabla\psi_{n}^{\ddagger}-\nabla\psi_{n}^{\dagger}\|_{\mathbb{P}_{n}}+\|\nabla\psi_{n}^{\ddagger}-\nabla\psi_{n}^{*}\|_{\mathbb{P}_{n}}\leq
\nonumber
\\
\sqrt{\frac{2\Delta}{\alpha_{n}}}\cdot\big(\sqrt{\frac{1}{\beta^{\ddagger}}}+\sqrt{\frac{1}{\lambda(\beta^{\ddagger})^{2}-\frac{\mathcal{B}^{\dagger}}{2}}}\big),
\end{eqnarray}
i.e.,
$$\mathbb{W}_{2}^{2}(\nabla\psi_{n}^{\dagger}\sharp\mathbb{P}_{n},\overline{\mathbb{P}})\leq \|\nabla\psi_{n}^{*}-\nabla\psi_{n}^{\dagger}\|_{\mathbb{P}_{n}}^{2}\leq \frac{2\Delta}{\alpha_{n}}\cdot\big(\sqrt{\frac{1}{\beta^{\ddagger}}}+\sqrt{\frac{1}{\lambda(\beta^{\ddagger})^{2}-\frac{\mathcal{B}^{\dagger}}{2}}}\big)^{2}=O(\Delta),$$
where the first inequality follows from \cite[Lemma A.2]{korotin2019wasserstein}.
% \lingxiao{cite (Korotin et al., 2019a, Lemma A.2) for the first inequality}
\end{proof}

\section{Experimental details and extra results}
\label{sec-extra}
In this section, we provide experimental details and additional results. In Subsection \ref{sec-exp-toy-extra}, we demonstrate qualitative results of computed barycenters in the 2-dimensional space. In Subsection \ref{sec-metrics}, we discuss used metrics in more detail. In Subsection \ref{sec-exp-params}, we list the used  hyperparameters of our method (C$\mathbb{W}_{2}$B) and methods [SC$\mathbb{W}_2$B], [CR$\mathbb{W}$B].

\subsection{Additional Toy Experiments in 2D}
\label{sec-exp-toy-extra}

We provide additional qualitative examples of computed barycenters of probability measures on $\mathbb{R}^{2}$.
% computed by C$\mathbb{W}_{2}$B (ours), [SC$\mathbb{W}_2$B] and [CR$\mathbb{W}$B] methods.

\begin{figure}[!h]
     \centering
     \begin{subfigure}[b]{0.31\columnwidth}
         \centering
         \includegraphics[width=\textwidth]{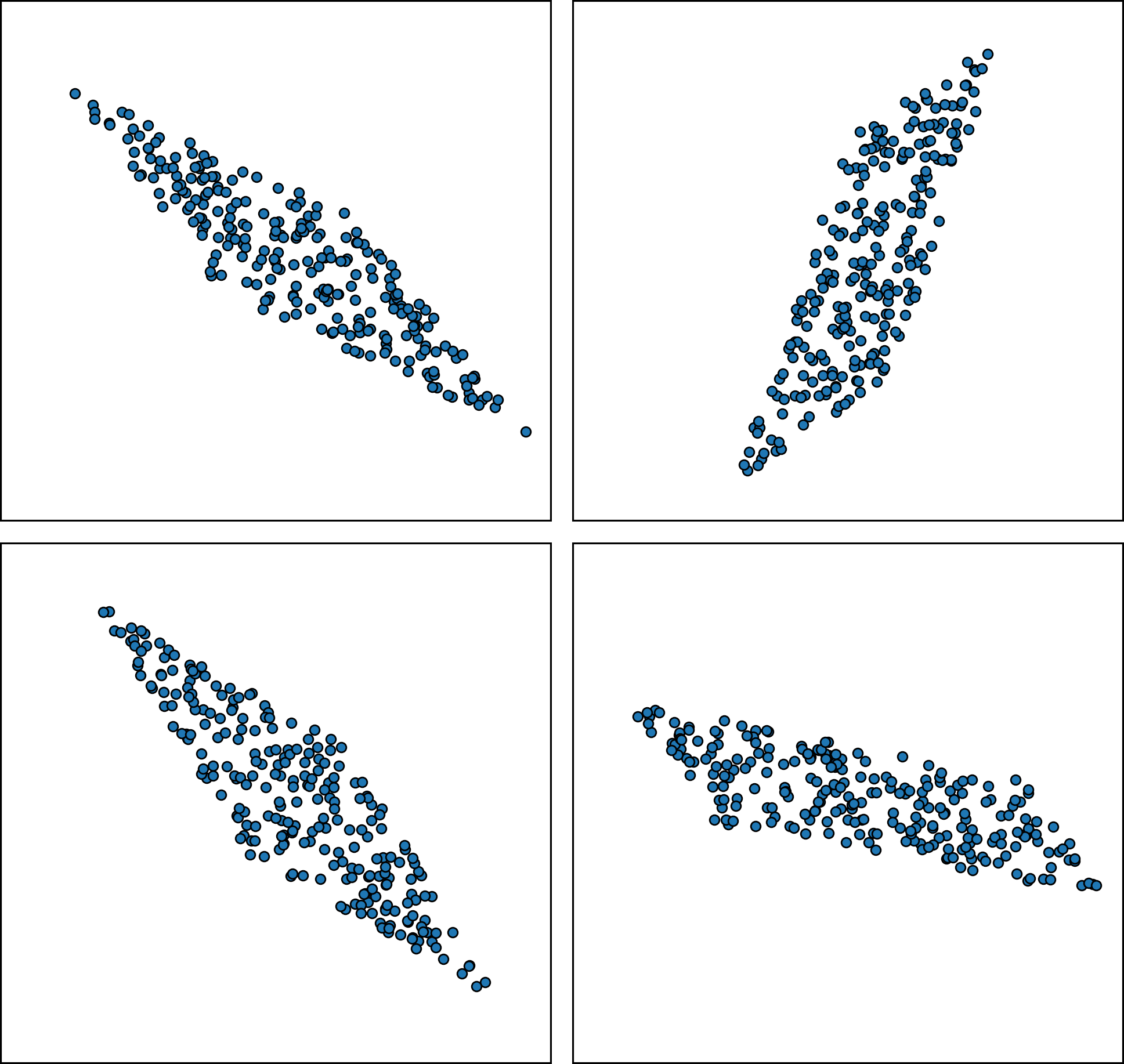}
         \caption{Input distributions $\{\mathbb{P}_{n}\}$}
     \end{subfigure}
     \hfill
     \begin{subfigure}[b]{0.31\columnwidth}
         \centering
         \includegraphics[width=\linewidth]{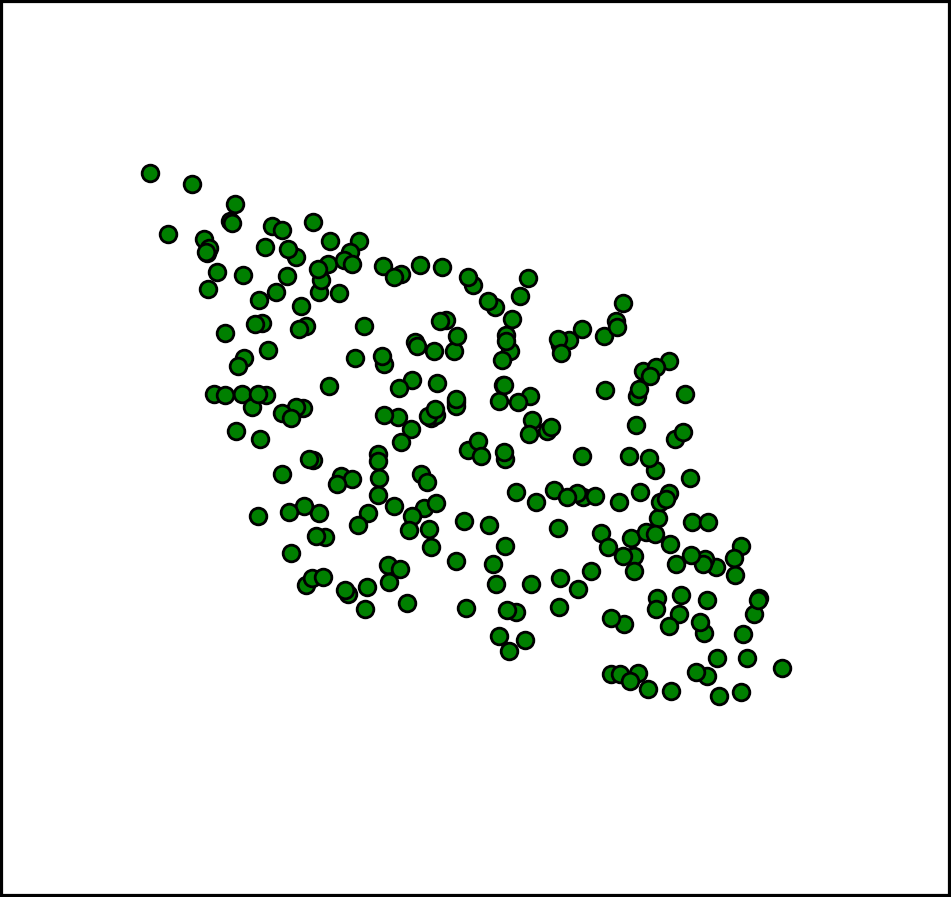}
     \caption{True barycenter $\overline{\mathbb{P}}$}
         \label{fig:ls-dim2-scwb-bar}
     \end{subfigure}
     \hfill
     \begin{subfigure}[b]{0.31\columnwidth}
         \centering
         \includegraphics[width=\linewidth]{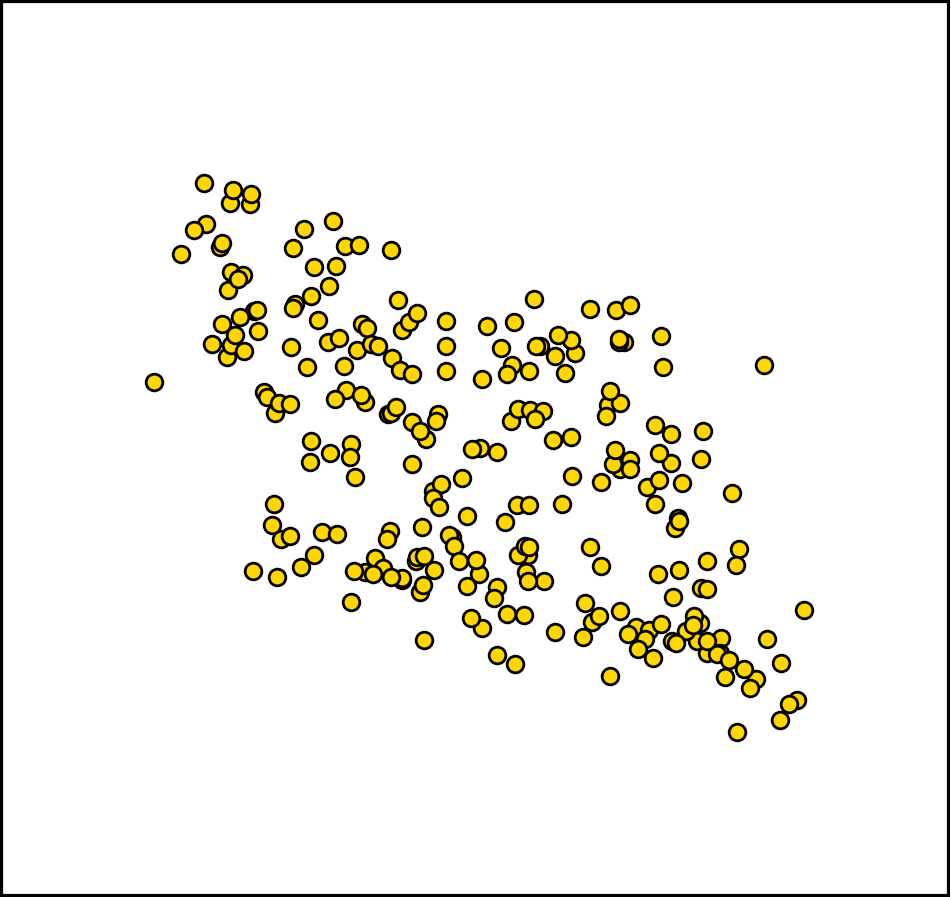}
         \caption{SC$\mathbb{W}_2$B, generated distribution $g\sharp\mathbb{S}$}
         \label{fig:ls-dim2-scwb-gen}
     \end{subfigure}
     
     \vspace{3mm}
     \begin{subfigure}[b]{0.31\columnwidth}
         \centering
         \includegraphics[width=\linewidth]{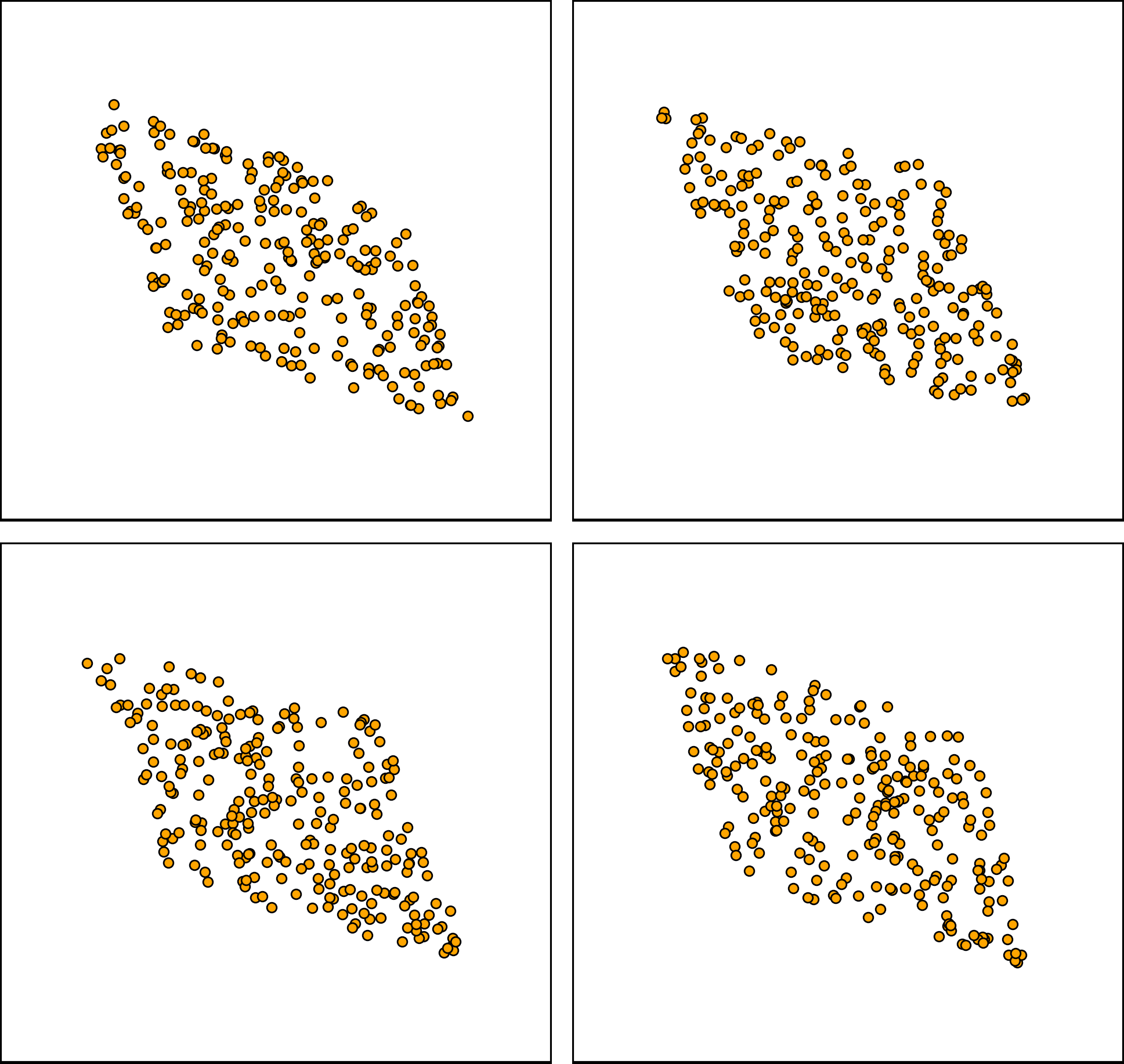}
         \caption{SC$\mathbb{W}_2$B, distributions $\nabla\psi_{n}^{\dagger}\sharp\mathbb{P}_{n}$}
     \end{subfigure}
     \hfill
     \begin{subfigure}[b]{0.31\columnwidth}
         \centering
         \includegraphics[width=\linewidth]{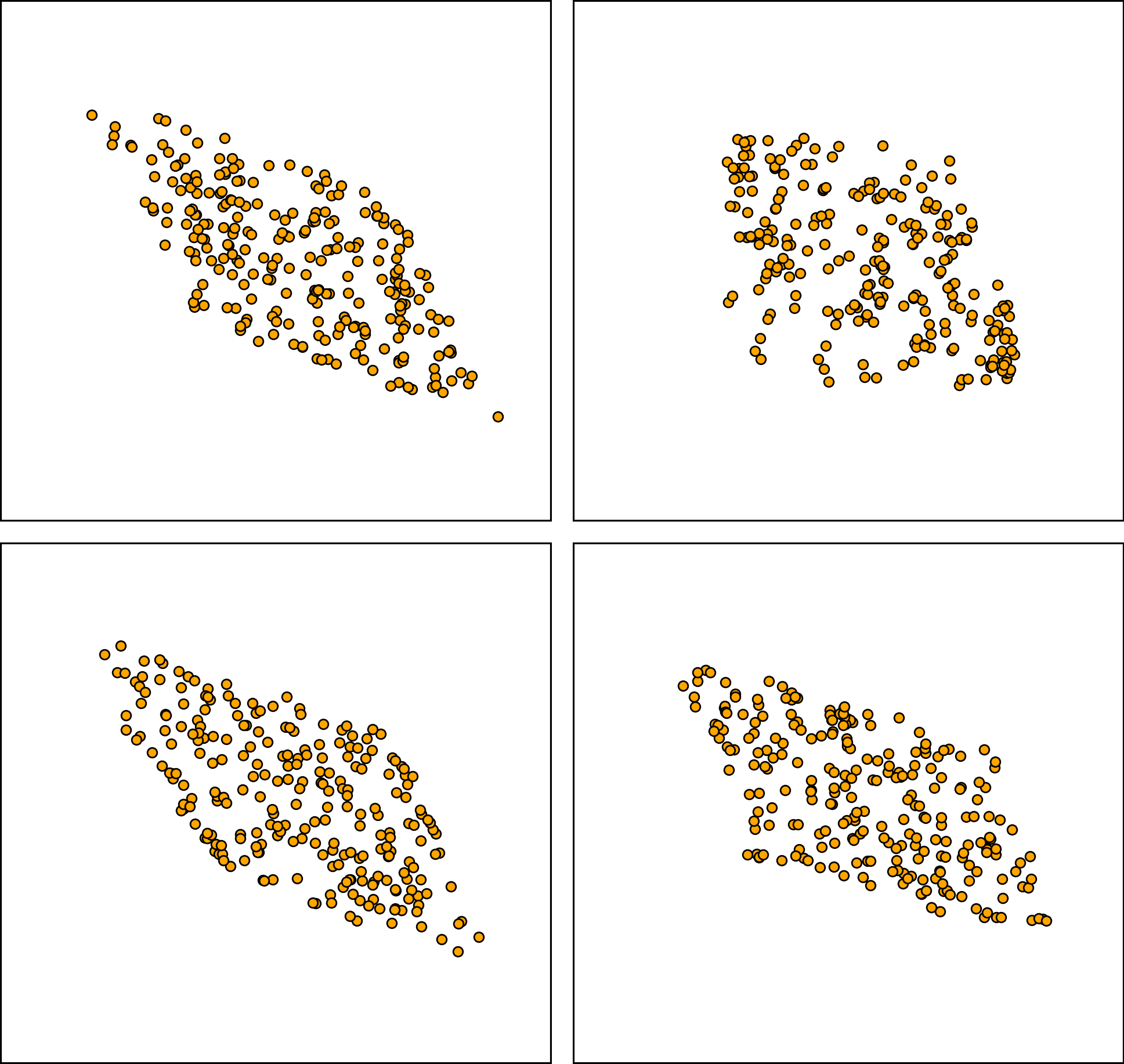}
         \caption{CR$\mathbb{W}$B, distributions ${\nabla\psi_{n}^{\dagger}\sharp\mathbb{P}_{n}}$}
         \label{fig:ls-dim2-crwb-pot}
     \end{subfigure}
     \hfill
     \begin{subfigure}[b]{0.31\columnwidth}
         \centering
         \includegraphics[width=\linewidth]{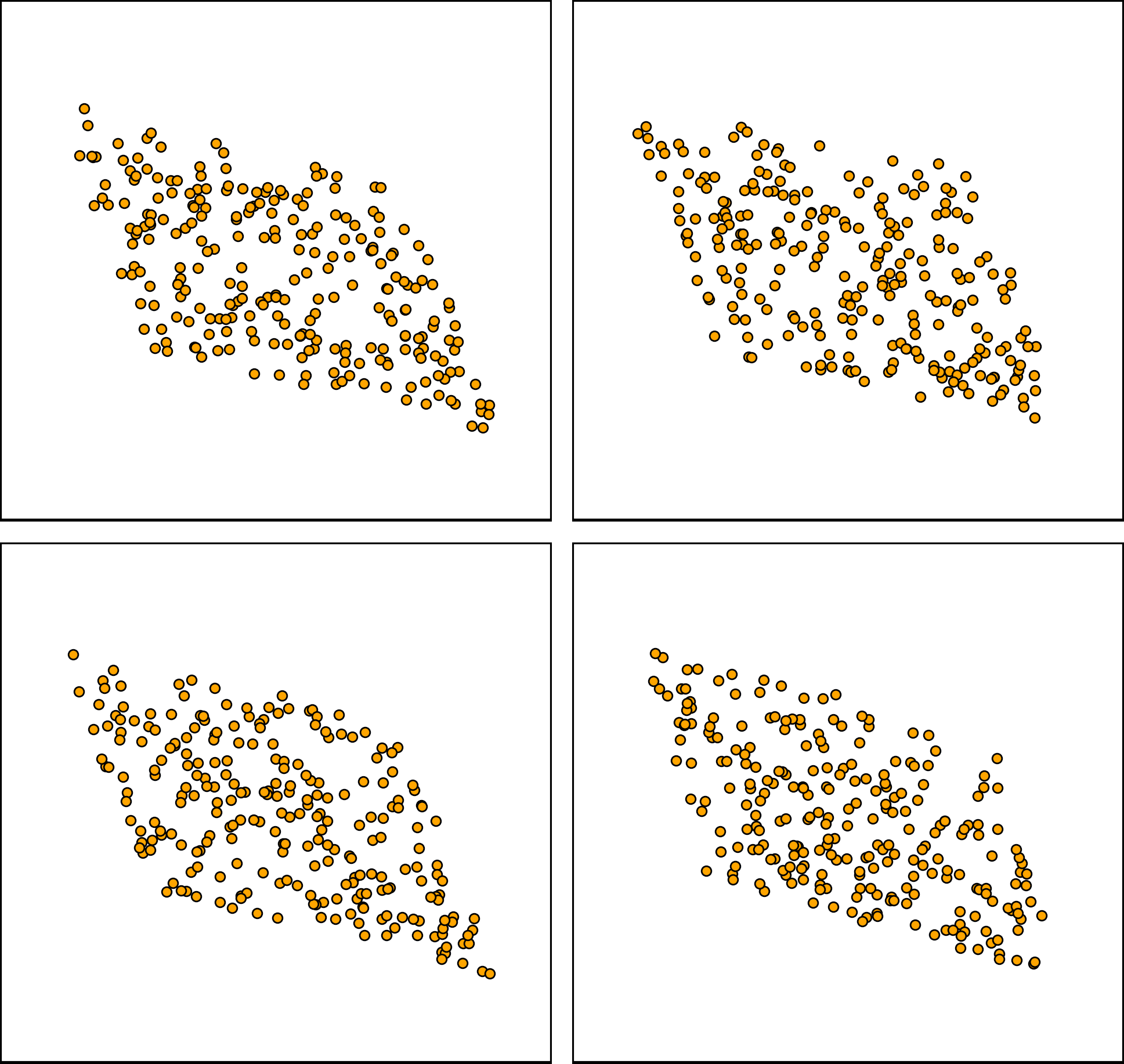}
         \caption{C$\mathbb{W}_{2}$B, distributions ${\nabla\psi_{n}^{\dagger}\sharp\mathbb{P}_{n}}$}
     \end{subfigure}
    \caption{Barycenter of a random location-scatter population computed by different methods.}
    \label{fig:ls-dim2-barycenters}
\end{figure}

In Figure \ref{fig:ls-dim2-barycenters}, we consider the location-scatter family $\mathcal{F}(\mathbb{P}_{0})$ with $\mathbb{P}_{0}=\text{Uniform}[-\sqrt{3},\sqrt{3}]^{D}$. 
In principle, all the methods capture the true barycenter. 
However, the generated distribution $g\sharp \mathbb{S}$ of [SC$\mathbb{W}_2$B] (Figure \ref{fig:ls-dim2-scwb-gen}) provides samples that lies outside of the actual barycenter's support (Figure \ref{fig:ls-dim2-scwb-bar}). 
Also, in [CR$\mathbb{W}$B] method, one of the potentials' pushforward measure (top-right in Figure \ref{fig:ls-dim2-crwb-pot}) has visual artifacts.

\begin{figure}[!h]
     \centering
     \begin{subfigure}[b]{0.38\columnwidth}
         \centering
         \includegraphics[width=\textwidth]{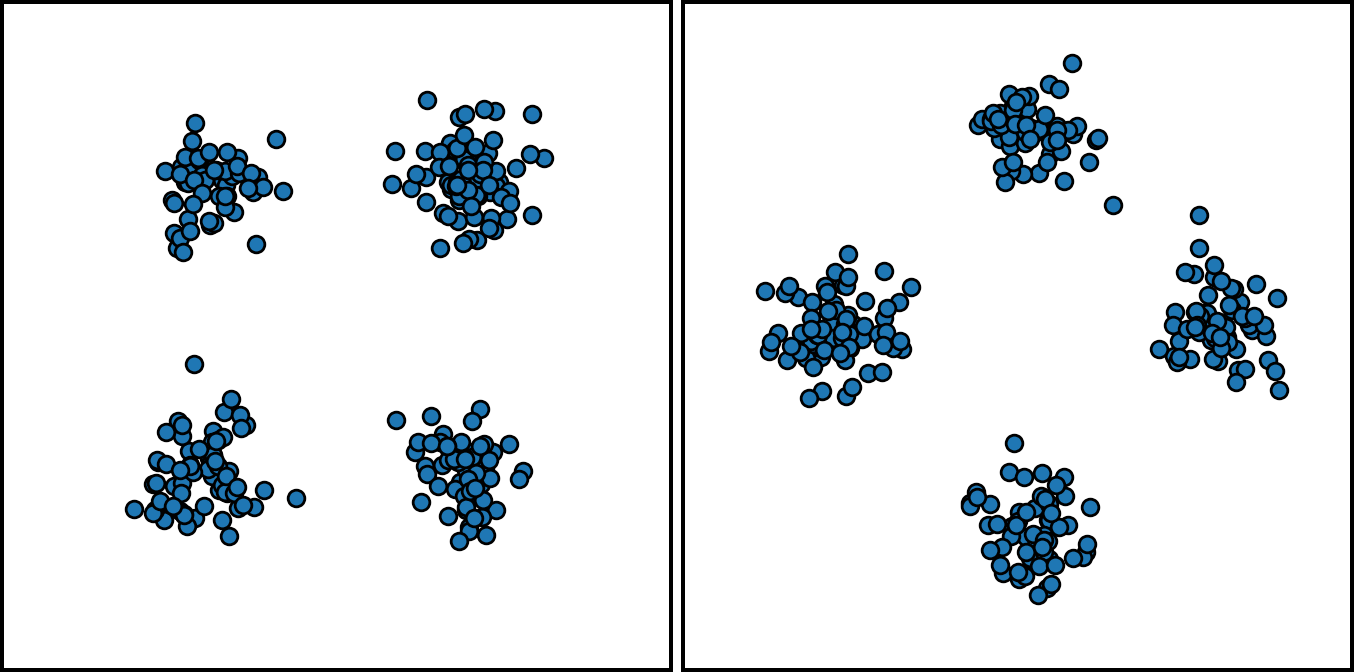}
         \caption{Inputs $\{\mathbb{P}_{n}\}$}
     \end{subfigure}
     \hfill
     \begin{subfigure}[b]{0.19\columnwidth}
         \centering
         \includegraphics[width=\linewidth]{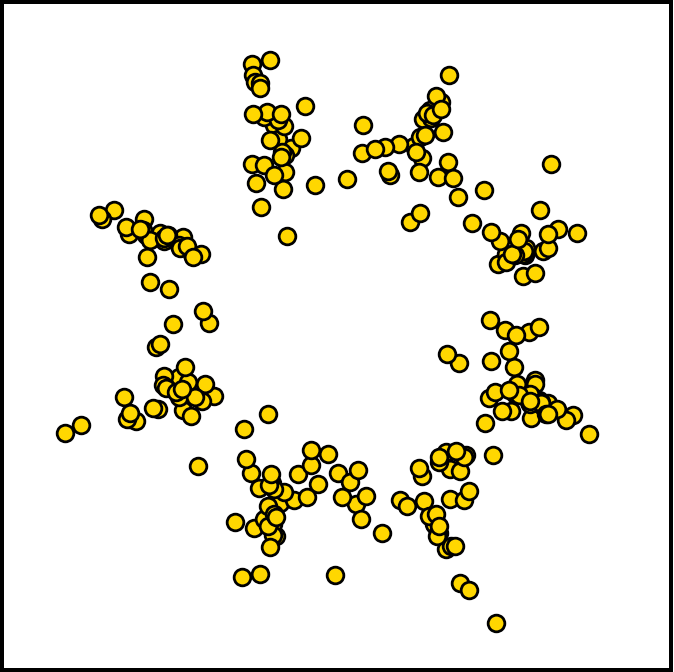}
         \caption{SC$\mathbb{W}_2$B $g\sharp\mathbb{S}$}
         \label{fig:gaumix-barycenters-gen}
     \end{subfigure}
     \hfill
     \vspace{2mm}
     \begin{subfigure}[b]{0.38\columnwidth}
         \centering
         \includegraphics[width=\linewidth]{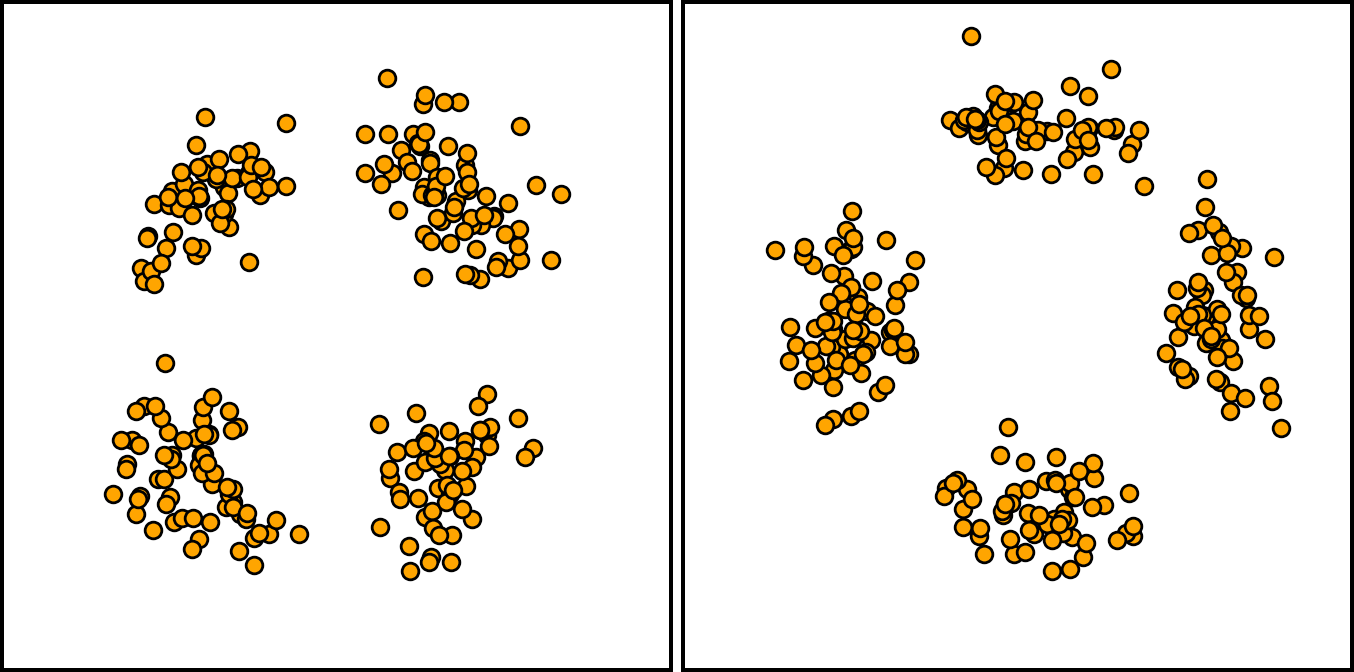}
         \caption{SC$\mathbb{W}_2$B $\nabla\psi_{n}^{\dagger}\sharp\mathbb{P}_{n}$}
     \end{subfigure}
     \hfill
     \begin{subfigure}[b]{0.38\columnwidth}
         \centering
         \includegraphics[width=\linewidth]{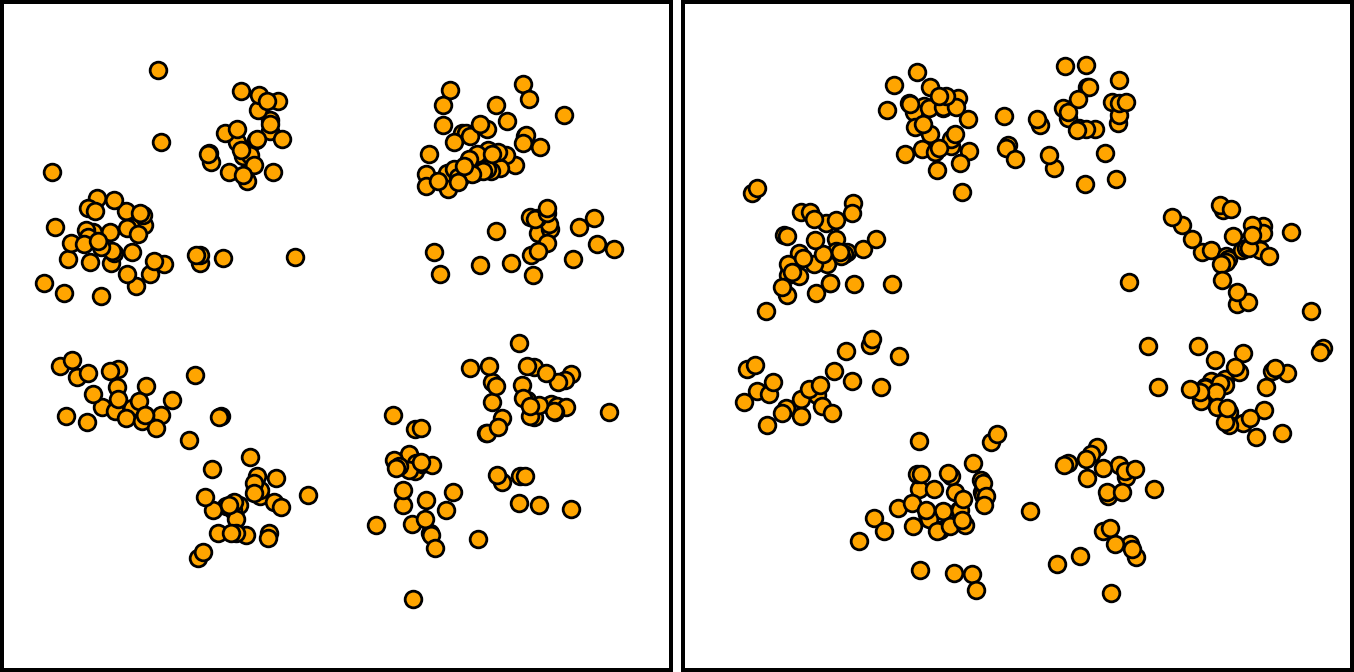}
         \caption{CR$\mathbb{W}$B ${\nabla\psi_{n}^{\dagger}\sharp\mathbb{P}_{n}}$}
     \end{subfigure}
     \hspace{10mm}
     \begin{subfigure}[b]{0.38\columnwidth}
         \centering
         \includegraphics[width=\linewidth]{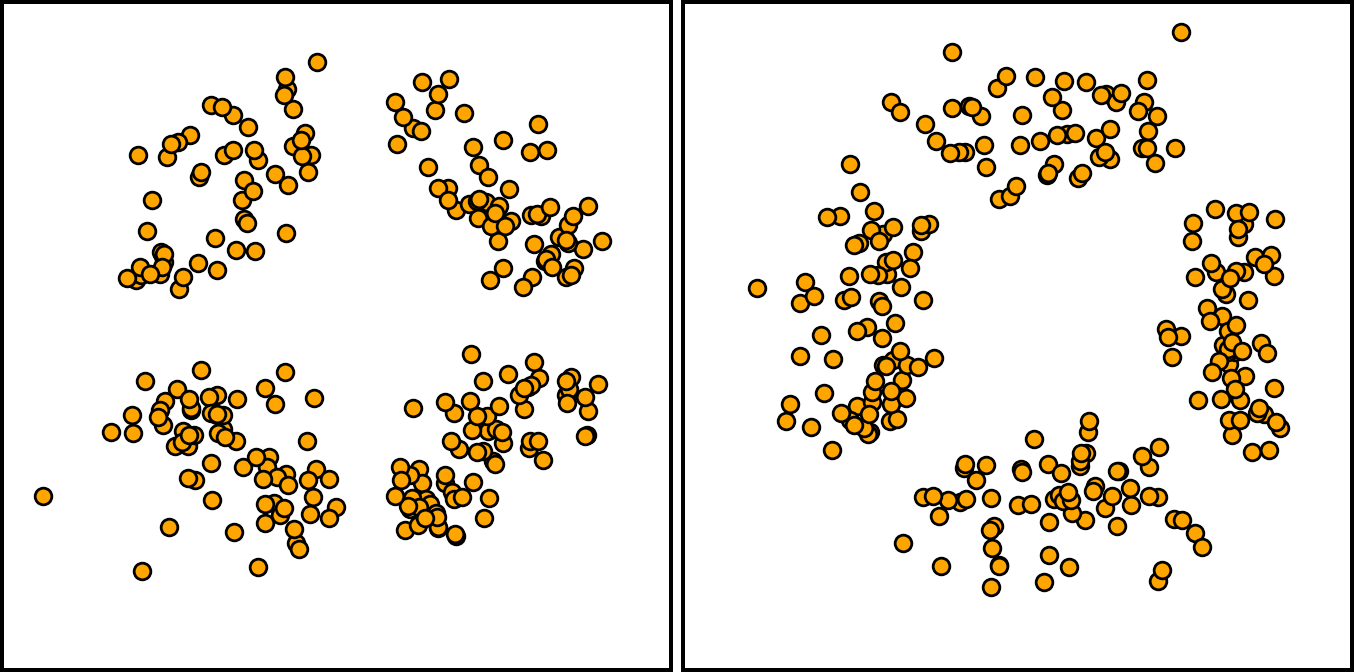}
         \caption{C$\mathbb{W}_{2}$B ${\nabla\psi_{n}^{\dagger}\sharp\mathbb{P}_{n}}$}
     \end{subfigure}
    \caption{Barycenter of a two 2D Gaussian mixtures.}
    \label{fig:gaumix-barycenters}
\end{figure}

In Figure \ref{fig:gaumix-barycenters}, we consider the Gaussian Mixture example by \citep{fan2020scalable}. The barycenter computed by [SC$\mathbb{W}_2$B] method (Figure \ref{fig:gaumix-barycenters-gen}) suffers from the behavior similar to mode collapse.

% \begin{figure}[!h]
%      \centering
%      \begin{subfigure}[b]{0.38\columnwidth}
%          \centering
%          \includegraphics[width=\textwidth]{pics/two_crusts_inputs.png}
%          \caption{Inputs $\{\mathbb{P}_{n}\}$}
%      \end{subfigure}
%      \hfill
%      \begin{subfigure}[b]{0.19\columnwidth}
%          \centering
%          \includegraphics[width=\linewidth]{pics/two_crusts_scwb_gen.png}
%          \caption{SC$\mathbb{W}$B $g\circ\mathbb{S}$}
%      \end{subfigure}
%      \hfill
%      \vspace{2mm}
%      \begin{subfigure}[b]{0.38\columnwidth}
%          \centering
%          \includegraphics[width=\linewidth]{pics/two_crusts_scwb_potentials.png}
%          \caption{SC$\mathbb{W}$B $\nabla\psi_{n}^{\dagger}\circ\mathbb{P}_{n}$}
%      \end{subfigure}
%      \hfill
%      \begin{subfigure}[b]{0.38\columnwidth}
%          \centering
%          \includegraphics[width=\linewidth]{pics/two_crusts_cwb_potentials.png}
%          \caption{CR$\mathbb{W}$B ${\nabla\psi_{n}^{\dagger}\circ\mathbb{P}_{n}}$}
%      \end{subfigure}
%      \hspace{10mm}
%      \begin{subfigure}[b]{0.38\columnwidth}
%          \centering
%          \includegraphics[width=\linewidth]{pics/two_crusts_w2cb_potentials.png}
%          \caption{C$\mathbb{W}_{2}$B ${\nabla\psi_{n}^{\dagger}\circ\mathbb{P}_{n}}$}
%      \end{subfigure}
%     \caption{Barycenter of a two 2D Gaussian mixtures.}
%     \label{fig:two-crusts-barycenters}
% \end{figure}

\subsection{Metrics}
\label{sec-metrics}

The unexplained variance percentage (UVP) (introduced in Section \ref{sec-experiments}) is a natural and straightforward metric to assess the quality of the computed barycenter. However, it is difficult to compute in high dimensions: it requires computation of the Wasserstein-2 distance. Thus, we use different but highly related metrics $\mathcal{L}^{2}$-UVP and B$\mathbb{W}_{2}^{2}$-UVP.

To access the quality of the recovered potentials $\{\psi^{\dagger}_{n}\}$ we use $\mathcal{L}^{2}$-UVP defined in \eqref{l2-uvp-def}.
%\begin{equation}
%\mathcal{L}^{2}\mbox{-UVP}(\nabla\psi^{\dagger}_{n},\mathbb{P}_{n})=100\frac{\|\nabla\psi^{\dagger}_{n}-\nabla\psi^{*}_{n}\|_{\mathbb{P}_{n}}^{2}}{\mbox{\small Var}(\overline{\mathbb{P}})}\%\qquad\bigg[\geq 100\frac{\mathbb{W}_{2}^{2}\big(\nabla\psi^{\dagger}_{n}\circ\mathbb{P}_{n}, \overline{\mathbb{P}}\big)}{\frac{1}{2}\mbox{\small Var}(\overline{\mathbb{P}})}\%\bigg].
%\label{l2-uvp-def}
%\end{equation}
$\mathcal{L}^{2}$-UVP compares not just pushforward distribution $\nabla\psi^{\dagger}_{n}\sharp\mathbb{P}_{n}$ with the barycenter $\overline{\mathbb{P}}$, but also the resulting transport map with the optimal transport map $\nabla\psi^{*}_{n}$. It bounds $\text{UVP}(\nabla\psi^{\dagger}_{n}\sharp\mathbb{P}_{n})$ from above, thanks to \citep[Lemma A.2]{korotin2019wasserstein}. Besides, $\mathcal{L}^{2}$-UVP naturally admits \emph{unbiased} Monte Carlo estimates using random samples from $\mathbb{P}_{n}$.

For measure-based optimization method, we also evaluate the quality of the generated measure $g\sharp\mathbb{S}$ using Bures-Wasserstein UVP defined in \eqref{bw-uvp-def}.
%\begin{equation}\text{B}\mathbb{W}_{2}^{2}\text{-UVP}(g\circ\mathbb{S})=100\frac{\text{B}\mathbb{W}_{2}^{2}(g\circ\mathbb{S},\overline{\mathbb{P}})}{\frac{1}{2}\mbox{\small Var}(\overline{\mathbb{P}})}\%\qquad\bigg[\leq 100\frac{\mathbb{W}_{2}^{2}\big(g\circ\mathbb{S}, \overline{\mathbb{P}}\big)}{\frac{1}{2}\mbox{\small Var}(\overline{\mathbb{P}})}\%\bigg]
%\label{bw-uvp-def}
%\end{equation}
 For measures $\mathbb{P},\mathbb{Q}$ whose covariance matrices are not degenerate, $\text{B}\mathbb{W}_{2}^{2}$ is given by
$$\text{B}\mathbb{W}_{2}^{2}(\mathbb{P},\mathbb{Q})=\frac{1}{2}\|\mu_{\mathbb{P}}-\mu_{\mathbb{Q}}\|^{2}+\big[\frac{1}{2}\Tr \Sigma_{\mathbb{P}}+\frac{1}{2}\Tr \Sigma_{\mathbb{Q}}-\Tr(\Sigma_{\mathbb{P}}^{\frac{1}{2}}\Sigma_{\mathbb{Q}}\Sigma_{\mathbb{P}}^{\frac{1}{2}})^{\frac{1}{2}}\big].$$
Bures-Wasserstein metric compares $\mathbb{P},\mathbb{Q}$ by considering only their first and second moments. It is known that $\text{B}\mathbb{W}_{2}^{2}(\mathbb{P},\mathbb{Q})$ is a lower bound for $\mathbb{W}_{2}^{2}(\mathbb{P},\mathbb{Q})$, see \citep{dowson1982frechet}. Thus, we have $\text{B}\mathbb{W}_{2}^{2}\text{-UVP}(g\sharp\mathbb{S})\leq \text{UVP}(g\sharp\mathbb{S})$. In practice, to compute $\text{B}\mathbb{W}_{2}^{2}\text{-UVP}(g\sharp\mathbb{S})$, we estimate means and covariance matrices of distributions by using $10^{5}$ random samples.

\subsection{Cycle Consistency and Congruence in Practice}
\label{sec-cycle-cong-check}

To assess the effect of the regularization of cycle consistency and the congruence condition in practice, we run the following sanity checks.

For cycle consistency, for each input distribution $\mathbb{P}_n$ we estimate (by drawing samples from $\mathbb{P}_n$) the value $\|\nabla \overline{\psi^{\ddagger}_{n}} \circ \nabla\psi^{\dagger}_{n}(x) - x\|_{\mathbb{P}_n}^{2} / \textup{Var}(\mathbb{P}_n)$. This metric can be viewed as an analog of the $\mathcal{L}^2$-UVP that we used for assessing the resulting transport maps. In all the experiments, this value does not exceed 2\%, which means that cycle consistency and hence conjugacy are satisfied well.

% This is further verified in the 2D case where we visualize the pushforward of samples from $\mathbb{P}_n$ by $\nabla \overline{\psi^{\ddagger}_{n}} \circ \nabla\psi^{\dagger}_{n}$; we obtain the original samples almost perfectly.

For the congruence condition, we need to check that $\sum_{n=1}^{N}\alpha_{n}\psi^{\dagger}_{n}(x)=\|x\|^2/2$. However, we do not know any straightforward metric to check this exact condition that is scaled properly by the variance of the distributions. Thus, we propose to use an alternative metric to check a slightly weaker condition on gradients, e.g., that $\sum_{n=1}^{N}\alpha_{n}\nabla\psi^{\dagger}_{n}(x)=x$. This is weaker due to the ambiguity of the additive constants. For this we can compute $\|\sum_{n=1}^{N}\alpha_{n}\nabla\psi^{\dagger}_{n}(x) - x\|_{\overline{\mathbb{P}}}^{2} / \textup{Var}(\overline{\mathbb{P}})$, where the denominator is the variance of the true barycenter. We computed this metric and found that it is also less than 2\% in all the cases, which means that congruence condition is mostly satisfied.

\subsection{Training Hyperparameters}
\label{sec-exp-params}
The code is written using the PyTorch framework. The networks are trained on a single GTX 1080Ti.
\subsubsection{Wasserstein-2 Continuous Barycenters (C$\mathbb{W}_{2}$B, our method)}
\label{sec-w2cb-params}
\textbf{Regularization}. We use $\tau=5$ and $\hat{\mathbb{P}}=\mathcal{N}(0,I_{D})$ in our congruence regularizer $\tau\cdot\mathcal{R}_{1}^{\hat{\mathbb{P}}}$. We use $\lambda=10$ for the cycle regularization $\lambda\cdot\mathcal{R}_{2}^{\mathbb{P}_{n}}$ for all $n=1,2,\dots, N$.

\textbf{Neural Networks (Potentials)}. To approximate potentials $\{\psi_{n}^{\dagger},\overline{\psi_{n}^{\ddagger}}\}$ in dimension $D$, we use $$\text{DenseICNN}[2;\max(64, 2D), \max(64, 2D), \max(32, D)]$$
with CELU activation function. DenseICNN is an input-convex dense architecture with additional convex quadratic skip connections. Here $2$ is the rank of each input-quadratic skip-connection's Hessian matrix. Each following number $\max(\cdot,\cdot)$ represents the size of a hidden dense layer in the sequantial part of the network. For detailed discussion of the architecture see \cite[Section B.2]{korotin2019wasserstein}. 
% For more details, see \citep{korotin2019wasserstein}.

\textbf{Training process}. We perform training according to Algorithm \ref{algorithm-main} of Appendix \ref{sec-algorithm-procedure}. We set batch size $K=1024$ and balancing coefficient $\gamma=0.2$. We use Adam optimizer by \citep{kingma2014adam} with a fixed learning rate $10^{-3}$. The total number of iterations is set to 50000.

\subsubsection{Scalable computation of Wasserstein Barycenters (SC$\mathbb{W}_2$B)}
\textbf{Generator Neural Network}. For the input noise distribution of the generative model we use ${\mathbb{S}=\mathcal{N}(0,I_{D})}$. 
For the generative network ${g:\mathbb{R}^{D}\rightarrow\mathbb{R}^{D}}$ we use a fully-connected sequential ReLU network with hidden layer sizes 
$${[\max(100,2D), \max(100,2D), \max(100,2D)]}.$$
Before the main optimization, we pre-train the network to satisfy $g(z)\approx z$ for all $z\in\mathbb{R}^{D}$. This has been empirically verified as a better option than random initialization of network's weights.

\textbf{Neural Networks (Potentials)}. We used exactly the same networks as in Subsection \ref{sec-w2cb-params}.

\textbf{Training process}. We perform training according to the min-max-min procedure described by \cite[Algorithm 1]{fan2020scalable}. The batch size is set to $1024$. We use Adam optimizer by \citep{kingma2014adam} with fixed learning rate $10^{-3}$ for potentials and $10^{-4}$ for generative network $g$. The number of iterations of the outer cycle (\emph{min}-max-min) number of iterations is set to 15000. Following \citep{fan2020scalable}, we use $10$ iterations per the middle cycle (min-\emph{max}-min) and $6$ iterations per the inner cycle (min-max-\emph{min}).

\subsubsection{Continuous Regularized Wasserstein Barycenters (CR$\mathbb{W}$B)}

\textbf{Regularization}. [CR$\mathbb{W}$B] method uses regularization to keep the potentials conjugate. The authors impose entropy or $\mathcal{L}^{2}$ regularization w.r.t. some proposal measure $\hat{\mathbb{P}}$; see \cite[Section 3]{li2020continuous} for more details. Following the source code provided by the authors, we use $\mathcal{L}_{2}$ regularization (empirically shown as a more stable option than entropic regularization). The regularization measure $\hat{\mathbb{P}}$ is set to be the uniform measure on a box containing the support of all the source distributions, estimated by sampling. The regularization parameter $\epsilon$ is set to $10^{-4}$.

\textbf{Neural Networks (Potentials)}. To approximate potentials $\{\psi_{n}^{\dagger},\overline{\psi_{n}^{\ddagger}}\}$ in dimension $D$, we use fully-connected sequential ReLU neural networks with layer sizes given by $$[\max(128, 4D), \max(128, 4D), \max(128, 4D)].$$
We have also tried using DenseICNN architecture, but did not experience any performance gain.

\textbf{Training process}. We perform training according to \cite[Algorithm 1]{li2020continuous}. We set batch size to $1024$. We use Adam optimizer by \citep{kingma2014adam} with fixed learning rate $10^{-3}$. The total number of iterations is set to 50000.

\end{document}